\documentclass[journal,hideappendix]{vgtc}        

\onlineid{1013}

\vgtccategory{Research}

\vgtcpapertype{algorithm/technique}

\newcommand\copyrightnoticeieee{%
\begin{tikzpicture}[remember picture,overlay]
\node[anchor=south,yshift=10pt] at (current page.south) {\fbox{\parbox{\dimexpr\textwidth-\fboxsep-\fboxrule\relax}{\footnotesize \textcopyright 2026 IEEE. Personal use of this material is permitted. Permission from IEEE must be obtained for all other uses, in any current or future media, including reprinting/republishing this material for advertising or promotional purposes, creating new collective works, for resale or redistribution to servers or lists, or reuse of any copyrighted component of this work in other works. DOI: }}};
\end{tikzpicture}%
}

\title{MultiCam: On-the-fly Multi-Camera Pose Estimation \\ Using Spatiotemporal Overlaps of Known Objects}
\def\approach{MultiCam\xspace}
\def\dataset{Femoral Nailing\xspace}
\author{%
  \authororcid{Shiyu Li}{0000-0002-0888-2496},
  \authororcid{Hannah Schieber}{0000-0002-5786-3283},
  \authororcid{Kristoffer Waldow}{0000-0002-5176-7530},\\
  \authororcid{Benjamin Busam }{0000-0002-0620-5774}, 
  \authororcid{Julian Kreimeier}{0000-0001-6861-711X}, and 
  \authororcid{Daniel Roth, \textit{Member, IEEE}}{0000-0001-5175-1566}
}

\authorfooter{
    \item
  	Shiyu Li, Hannah Schieber, Julian Kreimeier, and Daniel Roth are with Technische Universtität München, Human-Centered-Computing and Extended Reality Lab, Klinikum rechts der Isar, Orthopedics and Sports Orthopedics, Munich Institute of Robotics and Machine Intelligence (MIRMI), Germany. (e-mail: \{shiyu.li\,$|$\,hannah.schieber\,$|$\,julian.kreimeier\,$|$\,daniel.roth\}@tum.de).
  \item
    Kristoffer Waldow is with TH Köln, Computer Graphics Group and with Technische Universität München, Human-Centered-Computing and Extended Reality Lab, Germany. (e-mail: kristoffer.waldow@th-koeln.de).
  \item 
    Benjamin Busam is with Technische Universtität München, Photogrammetry and Remote Sensing Lab, Munich, Germany (e-mail: b.busam@tum.de).

}

\abstract{%
Multi-camera dynamic Augmented Reality (AR) applications require a camera pose estimation to leverage individual information from each camera in one common system. This can be achieved by combining contextual information, such as markers or objects, across multiple views. While commonly cameras are calibrated in an initial step or updated through the constant use of markers, another option is to leverage information already present in the scene, like known objects. Another downside of marker-based tracking is that  markers have to be tracked inside the field-of-view (FoV) of the cameras.

To overcome these limitations, we propose a constant dynamic camera pose estimation leveraging spatiotemporal FoV overlaps of known objects on the fly. To achieve that, we enhance the state-of-the-art object pose estimator to update our spatiotemporal scene graph, enabling a relation even among non-overlapping FoV cameras. To evaluate our approach, we introduce a multi-camera, multi-object pose estimation dataset with temporal FoV overlap, including static and dynamic cameras. Furthermore, in FoV overlapping scenarios, we outperform the state-of-the-art on the widely used YCB-V and T-LESS dataset in camera pose accuracy. Our performance on both previous and our proposed datasets validates the effectiveness of our marker-less approach for AR applications. 
  The code and dataset are available on \url{https://github.com/roth-hex-lab/IEEE-VR-2026-MultiCam}.
}


\teaser{
  \centering
  \includegraphics[width=\linewidth]{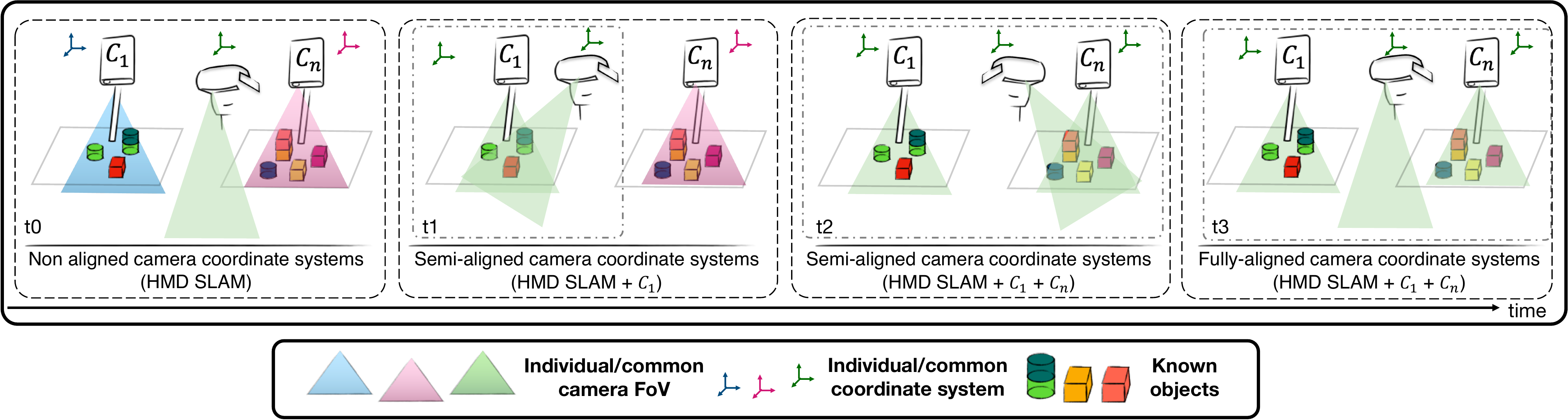}

\caption{Our proposed method \approach enhances the limited \ac{fov} of \ac{ar} \ac{hmd} by integrating more comprehensive spatial information from multiple external cameras with non-aligned camera coordinate systems (left figure).  We can dynamically calculate multiple camera poses using spatiotemporal object overlaps. If there are object overlaps between the \ac{hmd} and one external camera (second figure), we update the new calculated camera poses in the aligned coordinate system. In the end, the multi-camera system is fully aligned (right figure).}
  \label{fig:teaser}
}

\graphicspath{{figs/}{figures/}{pictures/}{images/}{./}} 

\usepackage{tabu}                      
\usepackage{booktabs}                  
\usepackage{lipsum}                    
\usepackage{mwe}                       

\usepackage{mathptmx}                  

\usepackage{times}                     
\usepackage[dvipsnames]{xcolor}
\usepackage{tabu}                      
\usepackage{booktabs}                  
\usepackage{lipsum}                    
\usepackage{mwe}                       
\usepackage[nolist]{acronym}

\usepackage{amsmath}
\usepackage{algorithm, algpseudocode}
\usepackage{tikz}

\usepackage{caption}
\usepackage{subcaption}
\usepackage{graphicx}
\usepackage{tabularx}
\usepackage{multirow}
\usepackage{tikz}
\usepackage{gensymb}
\usepackage{booktabs, makecell}
\usepackage{rotating}
\usepackage[export]{adjustbox}
\newcommand{\xmark}{%
\tikz[scale=0.23] {
    \draw[line width=0.7,line cap=round] (0,0) to [bend left=6] (1,1);
    \draw[line width=0.7,line cap=round] (0.2,0.95) to [bend right=3] (0.8,0.05);
}}
\newcommand{\cmark}{%
\tikz[scale=0.23] {
    \draw[line width=0.7,line cap=round] (0.25,0) to [bend left=10] (1,1);
    \draw[line width=0.8,line cap=round] (0,0.35) to [bend right=1] (0.23,0);
}}

\begin{document}
\begin{acronym}[Bspwwww.]  

\acro{ar}[AR]{Augmented Reality}
\acro{add}[ADD]{Average Distance Error}
\acro{add-auc}[ADD-AUC]{Average Distance Error under Curve}
\acro{ate}[ATE]{absolute trajectory error}
\acro{ba}[BA]{bundle adjustment}
\acro{bvip}[BVIP]{blind or visually impaired people}
\acro{cnn}[CNN]{convolutional neural network}
\acro{dlt}[DLT]{direct linear transformation}
\acro{dcc}[DCC]{dynamic coordinate classifier}
\acro{fov}[FoV]{field of view}
\acro{gan}[GAN]{generative adversarial network}
\acro{gcn}[GCN]{graph convolutional Network}
\acro{gnn}[GNN]{Graph Neural Network}
\acro{gbot}[GBOT]{Graph-based Object Tracking}
\acro{hmi}[HMI]{Human-Machine-Interaction}
\acro{hmd}[HMD]{head-mounted display}
\acro{dof}[DoF]{Degrees of Freedom}
\acro{mr}[MR]{mixed reality}
\acro{iot}[IoT]{internet of things}
\acro{icp}[ICP]{Iterative Closest Point}
\acro{llff}[LLFF]{Local Light Field Fusion}
\acro{bleff}[BLEFF]{Blender Forward Facing}

\acro{lpips}[LPIPS]{learned perceptual image patch similarity}
\acro{nerf}[NeRF]{neural radiance fields}
\acro{nvs}[NVS]{novel view synthesis}
\acro{mlp}[MLP]{multilayer perceptron}
\acro{mrs}[MRS]{Mixed Region Sampling}

\acro{or}[OR]{Operating Room}
\acro{pbr}[PBR]{physically based rendering}
\acro{psnr}[PSNR]{peak signal-to-noise ratio}
\acro{pnp}[PnP]{Perspective-n-Point}
\acro{roi}[ROI]{region-of-interest}
\acro{scenertm6d}[SceneRTM6D]{dynamic scene graph for real-time multi-view 6D pose estimation}
\acro{sus}[SUS]{system usability scale}
\acro{ssim}[SSIM]{similarity index measure}
\acro{sfm}[SfM]{structure from motion}
\acro{sift}[SIFT]{scale-invariant feature transform}
\acro{slam}[SLAM]{Simultaneous Localization and Mapping}
\acro{scg}[SSG]{Semantic Scene Graph}
\acro{tp}[TP]{True Positive}
\acro{tn}[TN]{True Negative}
\acro{thor}[thor]{The House Of inteRactions}
\acro{ueq}[UEQ]{User Experience Questionnaire}
\acro{vr}[VR]{Virtual Reality}
\acro{voom}[VOOM]{visual object odometry and mapping framework}
\acro{who}[WHO]{World Health Organization}
\acro{xr}[XR]{Extended reality}
\acro{ycb}[YCB]{Yale-CMU-Berkeley}
\acro{yolo}[YOLO]{you only look once}

\end{acronym}

\firstsection{Introduction}

\maketitle
\copyrightnoticeieee

\acf{ar} \acfp{hmd} are typically equipped with built-in cameras for environment sensing and object tracking. However, their dynamic \acf{fov} is inherently constrained to a narrow, egocentric perspective. In complex industrial and medical  scenarios, such a limited sensing capability can lead to the omission of critical contextual information from the surrounding environment \cite{ai2024calibration, song2022if}. To address this issue, static external cameras can be employed to extend the overall \ac{fov} of the application \cite{kleinbeck_artfm_2022}. But integrating additional cameras requires a pose estimation step to align their individual coordinate systems with that of the \ac{hmd}, enabling consistent joint sensing.

Dynamic cameras can update their pose by IMU data processing \cite{ungureanu2020hololens}, outside-in tracking \cite{barai2020outside}, and \ac{slam} approaches \cite{sun2020fast}. However, when IMU or \ac{slam} approaches are used alone for \ac{hmd}, they remain prone to accumulating errors over runtime, which requires their updated pose estimation or recalibration in relation to the static cameras~\cite{ai2024calibration, song2022if}, i.e., estimating and updating the camera pose. Fusing static and dynamic cameras requires estimating the pose of each camera by shared information in at least two static \ac{fov}s. This is usually implemented by optical markers \cite{fiala2005artag}  or calibration patterns \cite{zhang2002flexible}. However, having such markers is oftentimes not feasible in the application context where an updated pose is constantly needed due to dynamic scene parts. Furthermore, the implementation can be difficult in challenging environments such as sterile operating rooms. Firstly, the markers have to be sterilized, adding overhead in the medical workflow; secondly, awareness to always keep the markers in the \ac{fov} has to be raised. 

Another solution to address this problem could be the use of known objects as substitutes for markers, such as surgical instruments already present in the operating room. Similar conditions apply to industrial use cases, such as having known components during assembly. To leverage known objects as markers, each of their poses needs to be estimated. Several approaches (e.g. \cite{labbe2020cosypose, xiang_posecnn_2017,nguyen2024gigapose,labbe2022megapose,schieber_asdf_2024,li_gbot_2024}) exist to estimate object poses, and multiple datasets have been proposed \cite{xiang_posecnn_2017,jung_housecat6d_2023,drost2017introducing, wang_phocal_2022}. But existing datasets \cite{tremblay2018falling,hinterstoisser_model_2013,xiang_posecnn_2017,duffhauss2022mv6d,hodan_t-less_2017,drost2017introducing,tyree20226,liu2020keypose,wang_normalized_2019,hein2023next} do not cover multi-view setups with spatiotemporal overlaps between the cameras' \ac{fov}. However, combining this spatiotemporal overlap with object pose estimation can be an advantageous solution for dynamic camera pose estimation in dynamic scenes. To do so, an accurate and real-time capable object pose estimator needs to be applied in the scene. Thus, our approach requires the training of a 6D pose estimator. As a real-world dataset for every AR application is unrealistic and not scalable, synthetic data can be leveraged to meet training data demands\cite{denninger_blenderproc_2020, schieber_indoor_2024}. Additionally, synthetic data increases the scalability of our approach, as new objects can be easily added by expanding the training corpus \cite{denninger_blenderproc_2020, schieber_indoor_2024}.

\subsection{Contribution}

Current approaches and datasets do not address this spatiotemporal challenge in dynamic \ac{ar} camera setups. To address this gap, we a) overcome the need for dedicated optical markers and b) update static and dynamic cameras' pose data in a continuous, real-time manner with our approach (\approach). We leverage prior knowledge about existing objects in the temporarily shared fields of view and employ pose estimations of these objects. If we know objects in the scene, we can use the system on the fly without calibration. To summarize, we contribute: 

\begin{itemize}
\item A markerless multi-view camera pose estimation toolkit for RGB/RGB-D sensors using a real-time 6D object pose estimator
 
 \item A spatiotemporal scene graph fusing object pose information in temporarily-shared camera views and a novel object-level bundle adjustment for global optimization

\item A real-world multi-view multi-object benchmark dataset recorded with one \ac{ar} \ac{hmd} and two static cameras.
\end{itemize}

\autoref{fig:teaser} visualizes the application scenario of such a multi-camera setup. An \ac{hmd} user with a dynamic camera enters the scene. Our approach can handle static and dynamic multi-camera setups and update all cameras' pose data, based on temporarily shared \ac{fov} sensing. This allows \ac{ar} systems to extend the sensing \ac{fov} to multiple cameras and provides continuous context sensing information of objects, even when they are outside the current \ac{fov} of the \ac{hmd}.

\section{Related work}

\subsection{Camera Pose Estimation}

\subsubsection{Traditional Methods}
For the calibration of a single camera, Zhang \cite{zhang2002flexible} proposed a flexible calibration approach requiring the camera to observe a planar pattern shown at a few (at least two) different orientations. This approach builds the foundation for 3D computer vision. Fiala \cite{fiala2005artag} proposed ARTag, a fiducial marker system using digital techniques. Experimental results validate this system with a low error detection rate and high accuracy for \ac{ar}, robot navigation, and general applications between a camera and an object.

Most multi-camera calibration methods are marker-based\cite{ai2024calibration, guo2019online, huang2019research, rameau2022mc}. Rameau et al.\cite{rameau2022mc} introduced a toolbox for synchronized systems using multiple marker patterns, evaluating pose errors on synthetic data. However, their reliance on static markers limits use in \ac{ar} settings with space restrictions. Regarding \ac{ar} \ac{hmd}, Ai et al. \cite{ai2024calibration} proposed a robust virtual-to-real calibration that aligns the virtual and real spaces using retro-reflective markers. Song et al. \cite{song2022if} presented a marker-less calibration method integrating external cameras with \ac{ar} \ac{hmd} using point cloud registration, but the calibration of external cameras still relies on a fiducial marker.

 
\subsubsection{Learning-Based Methods}

Camera pose estimation determines a camera's position and orientation relative to a reference frame using semantic cues in a marker-less setup. It is crucial for tasks such as 6D pose estimation \cite{labbe2020cosypose,jung_housecat6d_2023}, novel view synthesis \cite{schieber2024nerftrinsic, lin2021barf}, and general scene understanding. Unlike camera calibration, which uses predefined patterns, pose estimation focuses solely on extrinsic parameters and can leverage object and background cues.

Scene coordinate regression methods \cite{kendall2015posenet, deng2022deep, shavit2021learning} estimate camera poses via deep learning in an end-to-end manner, eliminating the need for manual engineering or graph optimization. These methods can robustly re-localize the camera even when traditional point-based registration (e.g., SIFT) fails \cite{kendall2015posenet}. Roessle et al. \cite{roessle2023end2end} introduced an end-to-end pipeline based on feature matching and pose optimization, but their method suffers from scale ambiguity and does not consider dynamic camera scenarios.

\begin{table*}[t!]
\caption{\textbf{Dataset Overview.}  Comparison of existing instance-level 6D pose datasets with our proposed dataset in terms of camera setup, field-of-view (FOV) overlap, and object complexity.
}%
\resizebox{\textwidth}{!}{

\centering

\begin{tabular}{l | c c | c c | c c c| c c c c c | c c c }
\toprule
\multirow{2}{*}{Dataset} & 
        \multicolumn{2}{c|}{{\centering Modality}}& 
        \multicolumn{2}{c|}{{\centering Scene Type}}&
        \multicolumn{3}{c|}{{\centering Object}}&
        \multicolumn{5}{c|}{{\centering Camera}}&
        \multicolumn{3}{c}{{\centering Number}}\\ 
  &
  {\small RGB}          &
 {\small Depth}        &
  {\small Real}         &
  {\small Synthetic}   &
  {\small Cluttered}    &
  {\small Symmetry}     &
  {\small Transparent/}  &
    {\small Multi-view}   &
     {\small Static}   &
    {\small Dynamic}   &
    {\small XR }   &
     {\small Non-overlapping }   &
  {\small Categories}   &
  {\small Instances}      &
  {\small Scenes}     \\ 
  & 
    {\small }          &
 {\small }        &
  {\small }         &
  {\small }   &
  {\small }    &
  {\small }     &
  {\small Reflective}  &
    {\small }   &
     {\small }   &
    {\small }   &
    {\small  Device}   &
     {\small  FOV}   &
  {\small }   &
  {\small }      &
  {\small }     \\ 
\midrule
FAT~\cite{tremblay2018falling}                            & \cmark & \cmark &           & \cmark & \cmark & \cmark && \cmark&\cmark&\cmark&&& --     & $21$   & $>1$k \\
Linemod~\cite{hinterstoisser_model_2013} & \cmark & \cmark & \cmark    & \cmark & \cmark & \cmark  & & &&\cmark&&& --     & $15$   & $15$ \\
YCB~\cite{xiang_posecnn_2017}         & \cmark & \cmark & \cmark   & \cmark & \cmark & \cmark  & & &&\cmark&&& --     & $21$   & $92$ \\
MV-YCB~\cite{duffhauss2022mv6d}         & \cmark & \cmark &  & \cmark & \cmark &   \cmark  & & \cmark & \cmark & \cmark &&& --     &  21  & 8333 \\
T-LESS~\cite{hodan_t-less_2017}                                  & \cmark & \cmark & \cmark    & \cmark & \cmark &  \cmark  & &&&\cmark&&& --     & $30$   & $20$ \\
ITODD~\cite{drost2017introducing}                              &        & \cmark & \cmark   & \cmark & \cmark  & \cmark & &\cmark&&\cmark&&& --     & $28$   & $800$ \\
HOPE~\cite{tyree20226} & \cmark & \cmark & \cmark   & \cmark & \cmark & \cmark &  &\cmark&&\cmark& &  & -- & $28$   & $50$\\
TOD~\cite{liu2020keypose}                                 & \cmark & \cmark & \cmark   &    \cmark    & \cmark & \cmark    &  & \cmark &&\cmark&&& $3$    & $20$   & $10$ \\
Wet lab~\cite{hein2023next} & \cmark & \cmark  & \cmark   & \cmark &  &   & \cmark &\cmark&\cmark&\cmark&\cmark& & --    & 2   & 17(syn)/4(real)\\
\dataset (ours) & \cmark & \cmark  & \cmark   & \cmark & \cmark & \cmark  & \cmark &\cmark&\cmark&\cmark&\cmark&\cmark& --    & 9   & 400(syn)/2(real) \\ \bottomrule
\end{tabular}
}
\label{tab:dataset_comparison}%

\end{table*}

\subsection{6D Object Pose Estimation}

6D pose estimation predicts an object’s pose relative to the camera coordinate system. Single-view methods build the foundation for multi-view approaches; while single-network approaches offer faster inference, they often sacrifice accuracy~\cite{lu_rtmo_2023, hu2020single}. For \ac{ar} application, YOLO-based methods~\cite{schieber_asdf_2024,li_gbot_2024} are suitable due to a  good balance between speed and accuracy, but are limited to known objects. Approaches like~\cite{labbe2022megapose, nguyen2024gigapose}, which use 2D detectors such as CNOS~\cite{nguyen2023cnos}, are more generalizable for novel objects, but suffer from large runtime, which is far from real-time capability.

Multi-view techniques address depth ambiguities and occlusions~\cite{labbe2020cosypose, duffhauss2022mv6d, duffhauss2023symfm6d}. CosyPose~\cite{labbe2020cosypose} estimates both camera and object poses using RGB images and object-level RANSAC, but operates at 3 FPS, limiting AR applicability. MV6D\cite{duffhauss2022mv6d} and its extension SyMFM6D~\cite{duffhauss2023symfm6d} improve multi-view accuracy by assuming known camera poses and addressing object symmetry, achieving better results in cluttered scenes. However, both run below 10 FPS and require known camera poses, restricting their real-time use in \ac{ar}.



\subsection{Scene Graph Representation}

Scene graphs are beneficial for  scene understanding tasks requiring perception, interaction, and manipulation\cite{gay2019visual, rosinol20203d, wu2021scenegraphfusion}. \Acp{scg} \cite{wu2023incremental}  incorporate spatial relationships beyond object recognition. 

These methods primarily focus on estimating relationships across hierarchical levels, such as object~\cite{wald2020learning, nie2020total3dunderstanding},  and human-level semantics~\cite{rosinol20203d}. Wald et al.~\cite{wald2019rio} addressed object re-localization in dynamic environments, updating object poses accordingly. However, most existing approaches rely on monocular inputs~\cite{wald2020learning, wu2021scenegraphfusion} and lack integration of multi-camera data into the scene graph.

Spatio-temporal scene graphs~\cite{ji2020action, qiu2023virtualhome} have been proposed to capture dynamic interactions, particularly human-object relationships, supporting tasks like action recognition and video understanding. However, they typically rely on 2D object detection, omitting 3D cues such as object and human poses. In contrast, this work focuses on 6D pose estimation, leveraging scene graphs to connect objects and cameras within a unified 3D representation, facilitating more robust \ac{ar} applications.

 \subsection{6D Object Pose Datasets}

While single-camera 6D pose datasets~\cite{hinterstoisser_model_2013, xiang_posecnn_2017, hodan_t-less_2017} pose challenges such as occlusion, previous multi-view methods~\cite{labbe2020cosypose} evaluate multiple views on these datasets, using multiple frames from a single camera. To improve annotation accuracy, Li et al.\cite{li_bcot_2022} introduced the BCOT dataset with two static cameras observing dynamic objects, using multi-view geometry for pose optimization. Hein et al.~\cite{hein2023next} presented a dataset for surgical tool localization using five static RGB-D cameras and two \ac{ar} \ac{hmd}s; however, its shared FoV setup and limited object diversity make it unsuitable for non-overlapping, mobile multi-camera \ac{ar} scenarios.

As summarized in \autoref{tab:dataset_comparison}, we include reflective or metallic objects, which introduce challenging texture-less regions. We demonstrate it on a variety of challenging objects in medical use cases from the state-of-the-art household objects. Moreover, few datasets include both static and dynamic cameras, particularly \ac{xr} devices, under multi-view configurations. Our dataset addresses this gap by integrating static and dynamic viewpoints, incorporating materials that challenge conventional methods, and supporting both overlapping and non-overlapping FoV setups. Additionally, it targets the on-the-fly multi-camera pose estimation problem, aligning \ac{xr} device trajectories with external references under realistic \ac{ar} conditions without calibration.


\subsection{Research Gap}

Existing multi-view 6D object pose estimation methods (e.g.,\cite{labbe2020cosypose, duffhauss2022mv6d}) rely on static graphs and lack spatiotemporal reasoning for dynamic scene understanding, limiting their suitability for real-time \ac{xr} applications. Object-level bundle adjustment, jointly optimizing camera and object poses, has mainly been explored in single-view settings, often targeting re-projection error optimization, which are prone to detection errors and omit depth information. These methods generally neglect object feature properties and do not meet real-time constraints.

To address these limitations, we propose a spatiotemporal scene graph framework that processes RGB or RGB-D inputs and flexibly integrates external cameras and an \ac{ar} \ac{hmd} for real-time operation. Most existing datasets are single-view with limited pose diversity or multi-view datasets~\cite{li_bcot_2022, hein2023next} with limited camera setups. We contribute a new benchmark with both static and dynamic (\ac{ar} \ac{hmd}) camera setups to support robust multi-view pose evaluation.

\begin{figure*}[t!]
    \centering
    \includegraphics[width=\textwidth]{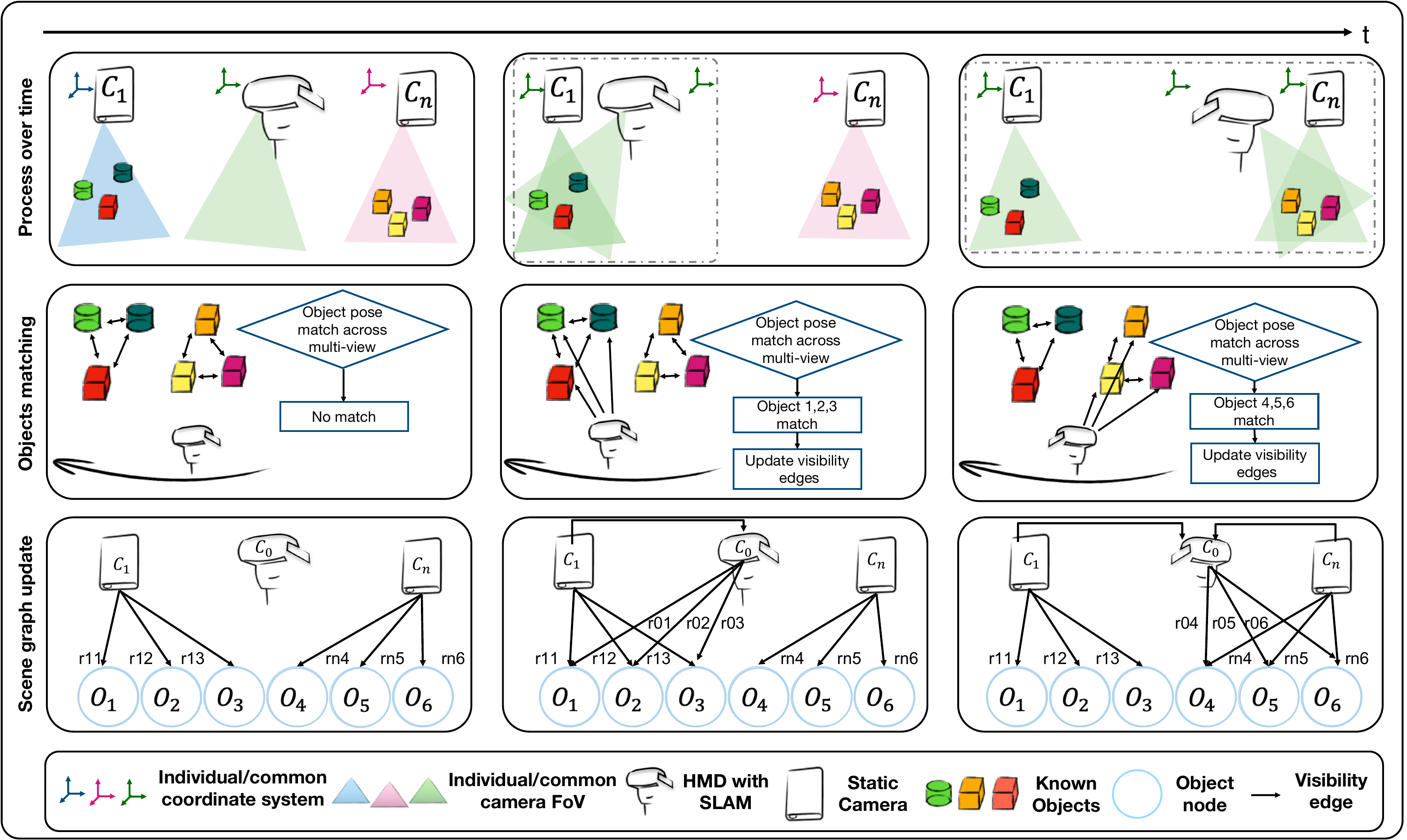}
     \caption{\textbf{Architecture of our proposed graph-based pose estimation.}  We generate object candidates from a set of captured RGB images using our 6D pose estimation module. Based on object candidates from different camera views, we select the matching candidates and calculate camera poses. Finally, we implement object and camera optimization using our novel spatiotemporal scene graph. We build our spatiotemporal scene graph using object information from different spaces and times. The icons represent camera nodes, i.e., \ac{hmd}, camera $C_1$, and camera $C_n$. The circles represent objects. If there are overlaps between two cameras, we calculate camera poses based on the \ac{hmd} SLAM coordinate system.}
    \label{fig:pose_architecture}
\end{figure*}

\section{Method}
To estimate dynamic camera poses from both \ac{ar} \ac{hmd}s and static cameras, our method combines scene graphs with multi-view 6D pose estimation. The pipeline begins with object pose candidates from YOLOPose. These are then integrated into our proposed spatiotemporal scene graph. Finally, we jointly optimize both camera and object poses using a pose-graph optimization framework.


\subsection{Symmetry-aware Keypoints Selection and 6D Object Pose Estimation}

We estimate 6D object poses by establishing 2D-3D correspondences using state-of-the-art keypoint-based methods~\cite{peng_pvnet_2019, hu2020single, wang_gdr-net_2021} from RGB inputs. Eight keypoints are sampled from the object’s 3D model via farthest point sampling (FPS) \cite{peng_pvnet_2019}, as 3D models, such as CAD models or scanned models, can be used for FPS. 

To enable real-time inference, we build a high-performance pose estimator on the YOLOX architecture\cite{yolox2021}. The network outputs bounding boxes and keypoints, which are processed by RANSAC PnP to recover the 6D pose. Our object detection head builds on RTM-O~\cite{lu_rtmo_2023}, a keypoint detector that leverages \ac{dcc} to enhance accuracy in one-stage detection. RTM-O dynamically assigns keypoint ranges based on predicted bounding boxes. Implementation details and hyperparameters are outlined in \autoref{subsec:impl}.

Symmetric objects often have repeated geometries that cause pose ambiguities. To address this, we define a set of valid symmetric transformations $\mathbf{S} = \{\mathbf{T}_{O_{S_1}}, \mathbf{T}_{O_{S_2}}, \hdots, \mathbf{T}_{O_{S_M}}\}$. Following~\cite{merrill_symmetry_2022}, we resolve symmetry by selecting keypoints corresponding to the pose closest to a predefined canonical view.

\subsection{Spatiotemporal Scene Graph}

We model dynamic scenes with cameras and objects by constructing a spatiotemporal scene graph comprising camera and object nodes. Given image sequence $I$ from multiple camera views, the scene graph predictor $SG$ generates all the cameras, objects, and connects each object with their relations to the camera in each frame, i.e., $SG: I \xrightarrow{} G$.  Each camera node is defined by a camera id and a camera pose, while the object node is defined by an instance id, a category id, and an object pose $T_{{C_a}{O_{a,\alpha}}}$, which represents the object pose of $\alpha$ observed by camera $a$. On each frame, the scene graph $SG = (O, C, R)$ consists of a set of objects $O = \{O_1, O_2, \dots\}$ that an object is visible by one or more cameras $C = \{C_1, C_2, \dots\}$, denoted by relationships of $R = \{\{r_{11}, r_{12}, \dots\}, \{r_{21}, r_{22}, \dots\}, \dots\}$, as shown in \autoref{fig:pose_architecture}. Here $r_{pq}$ denotes the relationship between camera $C_q$ and the visible object $O_p$. $r_{pq}$ equals to 1 if $O_p$ is visible by $C_q$, and 0 otherwise. 

Let's assume there are $\lvert C \rvert$ number of cameras and $\lvert O \rvert$ instances of objects in the scene. 
Similar to the pose graph, we assume there is a hierarchical structure between camera and object nodes. A visibility edge in the scene graph is defined as the connection between camera and object nodes. If objects are spatiotemporally visible in the camera view, then there is a connection $r_{pq}$ updated between the camera and object nodes.

\paragraph{Objects matching and initial camera pose estimation.} We use 6D object pose candidates from RGB images to calculate camera poses. To initialize the camera nodes with poses, we optimize the RANSAC algorithm proposed by ~\cite{labbe2020cosypose} using the unique camera pairs, which helps to reduce the overall runtime. In the case that camera $a$ observes object $\alpha$ and $\beta$ and camera b observes object $\gamma$ and $\delta$, where $\alpha \sim \gamma$ and $\beta \sim \delta$ ($\sim$ means the same category ids), we assume $\alpha$ and $\gamma$ as well as $\beta$ and $\delta$ are the same objects. Due to the pose ambiguity of symmetric objects, we use object pairs $(O_{a,\alpha}, O_{b,\beta})$ and
$(O_{a,\gamma},O_{b,\delta})$  to select the best initial camera pose using the equation from ~\cite{labbe2020cosypose} \begin{eqnarray}\label{eq:cab}
  {}^{C_a}T_{{C_b}} = {}^{C_a}T_{{O_{a,\alpha}}} S^\star {}^{C_b}T_{{O_{b,\beta}}}^{-1}
\end{eqnarray}
where ${}^{C_a}T_{{C_b}}$ is the pose of the camera b with respect to camera a, ${}^{C_a}T_{{O_{a,\alpha}}}$ is the object  $\alpha$ pose regard to camera a, ${}^{C_b}T_{{O_{b,\beta}}}$ is the object  $\beta$ pose regard to camera b,  $S^\star$ is the symmetry matrix among the symmetric transformations $S$, as discussed in the previous section, that minimizes their symmetric distance.

The camera pose between $C_a$ and $C_b$ can be estimated with enough inliers (at least three objects) when the matching error (symmetric distance) is smaller than a given threshold (5cm, roughly the object size). If the error is larger than the threshold, the objects will be classified as outliers. As both the camera and objects are movable, we need to define one camera with a known pose with respect to the world coordinate frame. 
 
 In \ac{ar} case, the front camera of the \ac{ar} \ac{hmd} has a known pose, since an \ac{ar} \ac{hmd} is equipped with sensors to track the camera poses~\cite{radkowski_hololens_2017}. The pose estimation of the external camera can be formulated as follows:

\begin{eqnarray}
   {}^{W}T_{C}(t) = {}^{W}T_{HMD}(t) {}^{HMD}T_{C}(t)
\end{eqnarray}

where ${}^{W}T_{C}(t)$ is the pose of external camera $C$ to be estimated in time t, ${}^{W}T_{HMD}(t)$ is the pose of \ac{hmd} regard to word coordinate system estimated by the inside-out tracking algorithm integrated in the \ac{hmd}, ${}^{HMD}T_{C}(t)$ is the pose of external camera $C$ regard to \ac{ar} \ac{hmd}, which can be calculated from object poses using \autoref{eq:cab}. 


\paragraph{Global Scene Aggregation.} To overcome the situation that all cameras need to share the same \ac{fov}, we  fuse the information from all camera views and propose a global scene aggregation based on our dynamic scene graph. Different from CosyPose ~\cite{labbe2020cosypose} that only selects the overlapping objects and removes the objects that are only visible to one camera, we keep all object candidates that are visible by all cameras. We combine the information and complete the scene information by aggregating the poses from different views.  If one object have more than one pose observed from multiple cameras, we use the pose from one camera view. We set the edge $r_{pq}$ to 0 if one object $O_p$ is regarded as an outlier by camera $C_q$ from the previous step. 

\paragraph{\ac{ar} \ac{hmd}-based spatiotemporal scene graph }


The camera pose estimation quality depends on the \ac{hmd} and external camera view. Therefore, it is crucial to select the camera view with high accuracy and drop any unreliable frames by the matching error. If the matching error is smaller than a threshold in the previous camera pose estimation step, we will update the camera poses. \autoref{fig:pose_architecture} shows one example of our spatiotemporal scene graph. In this example, we have one \ac{ar} \ac{hmd} (denoted as camera $C_0$) 
and two static cameras (denoted as camera $C_1$ and $C_n$). In the beginning, only the \ac{ar} \ac{hmd} pose is known, and the poses of the other two static cameras are unknown. In the first frames, there are no matching objects between these cameras. In the second keyframe, there is a match between \ac{hmd} $C_0$ and camera $C_1$. Therefore, we built an alignment between these two cameras. In the last keyframe, there are matching candidates for the other three objects between \ac{hmd} $C_0$ and camera $C_n$. All the cameras and objects are connected in our proposed spatiotemporal scene graph that fuses different camera views. The same process can be repeated if the number of cameras exceeds two.



\subsection{Object-level Bundle Adjustment}

We propose an optimization method based on object pose refinement using the spatiotemporal scene graph based on RGB or RGB-D images. We build our object-level bundle adjustment approach based on a probabilistic model ICG proposed by Stoiber et. al~\cite{stoiber_iterative_2022}. The probabilistic is defined using the log likelihood of the energy function. There are two modalities used to define the probabilistic model, i.e., region modality based on RGB image and depth modality based on depth image. 

The probabilistic model can be written as

\begin{equation}\label{eq:p30}
	p(\pmb{\theta}\mid \pmb{X}, \pmb{P}) = \mathcal{F}\big({}_\textrm{W}\pmb{X}(\pmb{\theta}_{cam}) - {}_\textrm{W}\pmb{P}(\pmb{\theta}_{obj})\big)
\end{equation}

where $\mathcal{F}$ is the function defined by region and depth modalities, $\pmb{X}$ are the 3D surface points observed by camera and $\pmb{P}$ are the 3D points from object models, $\pmb{\theta}_{cam}$ is the camera pose variation vector and $\pmb{\theta}_{obj}$ is the object pose variation vector. More details about pose variation can be found in ~\cite{stoiber_iterative_2022}. $W$ represents the world coordinate system. Therefore we have

\begin{equation}\label{eq:p30}
\begin{aligned}
	d p(\pmb{\theta}\mid \pmb{X}, \pmb{P}) &= \frac{\partial d_{\mathcal{F}}}{\partial {}_\textrm{W}\pmb{X}}
	\frac{\partial {}_\textrm{W}\pmb{X}}{\partial \pmb{\theta}_{cam}}
\bigg\vert_{\pmb{\theta}_{cam}=\pmb{0}} d \pmb{\theta}_{cam}\\
    &= -\frac{\partial d_{\mathcal{F}}}{\partial {}_\textrm{W}\pmb{P}}
	\frac{\partial {}_\textrm{W}\pmb{P}}{\partial \pmb{\theta}_{obj}}
	\bigg\vert_{\pmb{\theta}_{obj}=\pmb{0}} d \pmb{\theta}_{obj}
\end{aligned}
\end{equation}



which means that to optimize the camera poses, we need to move the camera during camera pose optimization in the opposite direction of object pose refinement. This conclusion is intuitive considering the following situation. One camera detects one object, and there is a positive offset in depth between the camera and the object. We should either move the object close to the camera or we should move the object back in the direction of the camera. The movements of the camera and the object are opposite. Therefore, we formulated the update for camera pose optimization with Gauss
Newton method based on object pose refinement as follows:

\begin{equation} \label{eq:m10}
	\pmb{\hat{\theta}_{cam}} = - \pmb{H}_{cam}^{-1}\pmb{g}_{cam},
\end{equation}

where the gradient vector $\pmb{g}_{cam}$ and the Hessian matrix $\pmb{H}_{cam}$ are the first- and second-order derivatives of the energy function $E(\pmb{\theta})$ or the negative logarithmic probability $-\ln(p(\pmb{\theta}))$. The update of camera poses equals the sum of items from the objects visible to this camera.

\begin{align} \label{eq:m21}
	\pmb{g}_\textrm{cam}^\top &= -\frac{\partial E_\textrm{cam}}{\partial \pmb{\theta}_\textrm{cam}}\bigg\vert_{\pmb{\theta}_\textrm{cam}=\pmb{0}}
	= -\sum_{i = 1}^n r_{ij}\frac{\partial E_i}{\partial \pmb{\theta}_i}\frac{\partial \pmb{\theta}_i}{\partial \pmb{\theta}_\textrm{cam}}\bigg\vert_{\pmb{\theta}_\textrm{cam}=\pmb{0}}
	= -\sum_{i = 1}^n r_{ij}\pmb{g}_i^\top \pmb{J}_i,\\[5pt]
	&\begin{aligned} \label{eq:m22}
		\pmb{H}_\textrm{cam} = \frac{\partial^2 E_\textrm{cam}}{\partial{\pmb{\theta}_\textrm{cam}}^{\!2}}\bigg\vert_{\pmb{\theta}_\textrm{cam}=\pmb{0}}
		&\approx \sum_{i = 1}^n r_{ij}\frac{\partial \pmb{\theta}_i}{\partial 	\pmb{\theta}_\textrm{cam}}^{\!\top} \frac{\partial^2 E_i}{\partial {\pmb{\theta}_i}^{\!2}} \frac{\partial \pmb{\theta}_i}{\partial \pmb{\theta}_\textrm{cam}}\bigg\vert_{\pmb{\theta}_\textrm{cam}=\pmb{0}}\\
		&\approx \sum_{i = 1}^n r_{ij}\pmb{J}_i^\top \pmb{H}_i\,\pmb{J}_i,
	\end{aligned}
\end{align}

with $\pmb{g}_i$ and $\pmb{H}_i$ the gradient vector and the Hessian matrix of the object $i$ in the scene graph, $r_{ij}$ is the previous mentioned relationship between camera j and object i in the scene graph and $\pmb{J}_i$ is the Jacobian matrix of camera.

The object-level bundle adjustment process is only implemented in the keyframes where there are overlapping objects between cameras. As the detected object poses may have large errors for some objects, we optimize the object poses and camera poses separately. In the case of multi-camera view and multiple objects, we assume the pose of one camera view is known, and both camera and object poses need to be optimized using the connection between camera and object nodes in a scene graph. For these objects visible by more than one camera and detected as inliers in the initial camera pose estimation process, the predefined energy function will be used to optimize both camera and object poses. For the objects which is only visible by one camera or detected as outliers, we only optimize the object poses.

\subsection{The \dataset Dataset}

\begin{figure}[t!]
    \centering\includegraphics[width=\columnwidth]{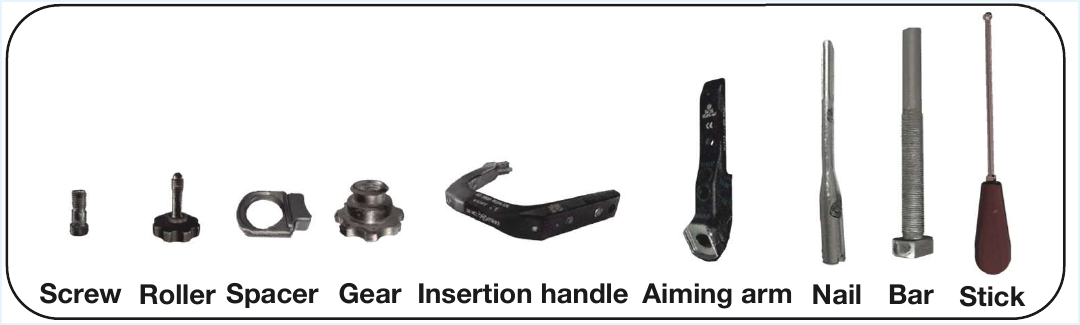}
    \caption{\textbf{Objects included in our \dataset dataset.}}
    \label{fig:scenertm6d_object}
\end{figure}

Currently, there is no usable multiple-view 6D pose dataset for \ac{ar} application. We create a small test dataset with nine parts of a surgical tool, as shown in \autoref{fig:scenertm6d_object}. These objects show different levels of pose estimation difficulty by varying object sizes, shapes, and textures.

\subsubsection{Training Set}

Our training set contains 10K synthetic images of reflective medical instruments. The images feature assembled and unassembled objects with 50\% of each variant. To improve domain randomization, we generated synthetic data using domain randomization~\cite{schieber_indoor_2024}. We apply different textures, scenes, overlays, distracting objects from T-less~\cite{hodan_t-less_2017}, varying backgrounds and light conditions. 

\subsubsection{Test Set}
We created a real-world test dataset using one HoloLens 2 \ac{hmd} and two Azure Kinects. We recorded the sequences and validated the camera poses using the OptiTrack system with 6 infrared cameras. Furthermore, we recorded camera poses from the HoloLens's \ac{slam} system. The images of the sequence are RGB images from the HoloLens' primary view with a resolution of 1280x720 and RGB-D images from two Azure Kinect DKs with a resolution of 1280x720.

Our dataset includes ground truth HoloLens coordinates captured by the OptiTrack system and the coordinates estimated by HoloLens' SLAM. We assume camera poses of HoloLens are known in the evaluation, and poses of other cameras are estimated based on the HoloLens camera. We use a TCP server to get the video stream from the HoloLens primary view. As there is time latency between different cameras, we carefully align the tracking poses to each camera based on object segmentations.

To compare the accuracy of the state-of-the-art camera pose estimation approaches with the mostly used marker-based camera calibration approaches, we also include a 5x4 Charuco calibration board with a 50mm checker and a 40 mm marker (in the same scale of object size) in the scene. To study the effects of distance on the pose accuracy, we recorded the scene in near distance (within 0.5 m) and far distance (0.75m - 1m), similar to ~\cite{hu_wide-depth-range_2021}.

\begin{table}[t!]
    \centering
    \caption{
        \label{tab:ycb_results_ADD}
        \textbf {Single-view object pose evaluation on YCB-V dataset ~\cite{xiang_posecnn_2017} with ADD(-S)-0.1d.}
        P.E. means whether the method is trained with 1 pose estimator for the whole dataset or 1 per object ($N$ objects in total). ($*$) denotes symmetric objects and ($-$) denotes unavailable results.}
    \resizebox{0.5\textwidth}{!}{
    \begin{tabular}{l | c c  c | cc } \toprule
        Approach &\multicolumn{3}{c|}{One-stage} &        \multicolumn{2}{c}{Two-stage} \\
        Method & PoseCNN & 
        SegDriven  &
        YOLOPose&
        \multicolumn{2}{c}{GDR-Net} \\ 
        & \cite{xiang_posecnn_2017} & 
        \cite{hu2019segmentation}  &
        (\textbf{Ours})&
        \multicolumn{2}{c}{\cite{wang_gdr-net_2021}} \\  \midrule
        P.E. & 1 & 1  & 1 & 1 & $N$ \\ \midrule
        002\_master\_chef\_can       &  3.6 & 33.0 & 70.2  & 51.7 & 41.5 \\
        003\_cracker\_box            & 25.1 & 44.6 & 85.4  & 45.1 & 83.2 \\
        004\_sugar\_box              & 40.3 & 75.6 & 73.8  & 83.9 & 91.5 \\
        005\_tomato\_soup\_can       & 25.5 & 40.8 & 73.7  & 48.3 & 65.9  \\
        006\_mustard\_bottle         & 61.9 & 70.6 & 92.0  & 92.2 & 90.2 \\
        007\_tuna\_fish\_can         & 11.4 & 18.1 & 60.2  & 29.1 & 44.2 \\
        008\_pudding\_box            & 14.5 & 12.2 & 82.0  & 39.7 & 2.8 \\
        009\_gelatin\_box            & 12.1 & 59.4 & 62.7  & 34.6 & 61.7 \\
        010\_potted\_meat\_can       & 18.9 & 33.3 & 70.5  & 36.3 & 64.9 \\
        011\_banana                  & 30.3 & 16.6 & 47.1  & 60.2 & 64.1 \\
        019\_pitcher\_base           & 15.6 & 90.0 & 100.0  & 96.3 & 99.0 \\
        021\_bleach\_cleanser        & 21.2 & 70.9 & 76.2  & 73.0 & 73.8 \\
        024\_bowl$^*$                & 12.1 & 30.5 & 24.9  & 35.0 & 37.7  \\
        025\_mug                     &  5.2 & 40.7 & 62.8  & 39.3 & 61.5  \\
        035\_power\_drill            & 29.9 & 63.5 & 90.5  & 57.7 & 78.5  \\
        036\_wood\_block$^*$         & 10.7 & 27.7 & 78.6  & 50.8 & 59.5  \\
        037\_scissors                &  2.2 & 17.1 & 73.5  & 6.6  & 3.9  \\
        040\_large\_marker           &  3.4 &  4.8 & 10.4  & 13.7 & 7.4  \\
        051\_large\_clamp$^*$        & 28.5 & 25.6 & 71.4  & 40.3 & 69.8  \\
        052\_extra\_large\_clamp$^*$ & 19.6 &  8.8 & 91.7  & 35.3 & 90.0  \\
        061\_foam\_brick$^*$         & 54.5 & 34.7 & 69.2  & 61.1 & 71.9 \\
        \midrule
        MEAN                         & 21.3 & 39.0  & \textbf{69.9} & 49.1 & 60.1 \\ \bottomrule
    \end{tabular}
    }  
\end{table}

\section{Evaluation}
We evaluate the proposed methods on the YCB-Video dataset for general multi-camera pose estimation and our \dataset dataset to show the \ac{ar} suitability.

\subsection{Metrics}

To evaluate the 6D object pose accuracy, we follow the state-of-the-art and report \ac{add} / \ac{add}-S~\cite{hinterstoisser_model_2013} and \ac{add-auc} ~\cite{xiang_posecnn_2017}. 

To evaluate camera poses, we considered two different errors, i.e., pose estimation errors and sensor drift errors. For pose estimation errors, we compare the estimated camera poses and ground truth camera poses~\cite{shavit2019introduction} in the predefined world coordinate. The state-of-the-art assumes constant overlapping \ac{fov} from multiple cameras, while our approach assumes temporal overlaps from static cameras with dynamic cameras, so we only compare the results in the overlapping case. Sensor drift errors are analyzed when \ac{ar} \ac{hmd} \ac{slam} is used. In our recorded dataset, we compared HoloLens tracking data and ground truth camera poses with a motion capture system.

\subsection{Implementation Details}
\label{subsec:impl}


The total loss for the proposed pose estimation network is 

\begin{equation}
\mathcal{L}=\lambda_{1}\mathcal{L}_{\textit{bbox}}+\lambda_{2}\mathcal{L}_{\textit{2D}}+\lambda_{3}\mathcal{L}_{\textit{3D}}+\lambda_{4}\mathcal{L}_{\textit{proxy}}+\lambda_{5}\mathcal{L}_{\textit{cls}}+\mathcal{L}_{\textit{vis}},\nonumber
\end{equation}
with hyperparameters $\lambda_{1},\lambda_{2},\lambda_{3},\lambda_{4}$ and $\lambda_{5}$ set at $\lambda_{1}=\lambda_{2}=5$, $\lambda_{3}=\lambda_{4}=10$, and $\lambda_{5}=2$, where $\mathcal{L}_{\textit{bbox}}$ is bounding box loss, $\mathcal{L}_{\textit{2D}}$ is the 2D MLE keypoint loss, $\mathcal{L}_{\textit{3D}}$ is the 3D keypoint loss proposed by \cite{hu_wide-depth-range_2021}, $\mathcal{L}_{\textit{proxy}}$ is the proxy loss in SimOTA for positive grid selection, $\mathcal{L}_{\textit{cls}}$ is object instance classification loss, represented
by varifocal loss and $\mathcal{L}_{\textit{vis}}$ is BCE loss for visibility. These hyperparameters are selected following \cite{lu_rtmo_2023} to keep a balance between these losses. 

We apply data augmentation, like flipping, affine transformation, which are not applicable for 3D keypoint loss. Therefore, we combine 2D and 3D keypoint loss to train our models in multiple stages.

We divide our training into three stages. Following the training steps of~\cite{lu_rtmo_2023}, the first involves training both the proxy branch and \ac{dcc} using 2D keypoint annotations, and the second shifts the target of the proxy branch to the decoded pose from DCC. In the last stage, we use no data augmentation and add a 3D keypoint loss.

\subsection{Experimental Setup}

We use a workstation with an Intel(R) Core(TM) i9-10980XE CPU and NVIDIA GeForce RTX 4090 GPU with 24GB VRAM. Our training batch size is set to 8 and the inference batch size is set to the number of camera views. Our object pose estimation algorithm extends YOLOX, is implemented in PyTorch for 6D pose estimation, and uses ONNX runtime to accelerate and optimize inference performance. Inference engine, our pose estimation, and the TCP server for HoloLens stream~\cite{dibene2022hololens} are implemented in C++ 17. The code of camera API (HoloLens and Azure Kinect), pose estimation is publicly available on GitHub to ensure reproducibility and re-usability for further future work.

\subsection{Evaluation on YCB-V and T-LESS Datasets}

\autoref{tab:ycb_results_ADD} shows the single-view evaluation results on the YCB-V dataset. Our enhanced one-stage approach outperformed the state-of-the-art methods in accuracy. We achieved an average ADD(S)-0.1d of 69.9 of all the objects.

\begin{figure*}[t!]
\setlength\tabcolsep{1pt}
\centering
\begin{tabularx}{\textwidth}{l XXXXX }
& \centering View 1 & \centering View 2 & \centering View 3 & \centering View 4 & \centering View 5 \cr \\
\rothead{Ground truth}        &   \includegraphics[width=\hsize,valign=m]{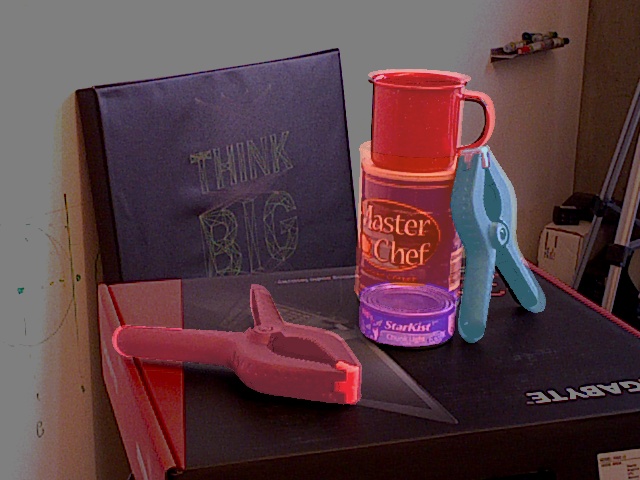}
                        &   \includegraphics[width=\hsize,valign=m]{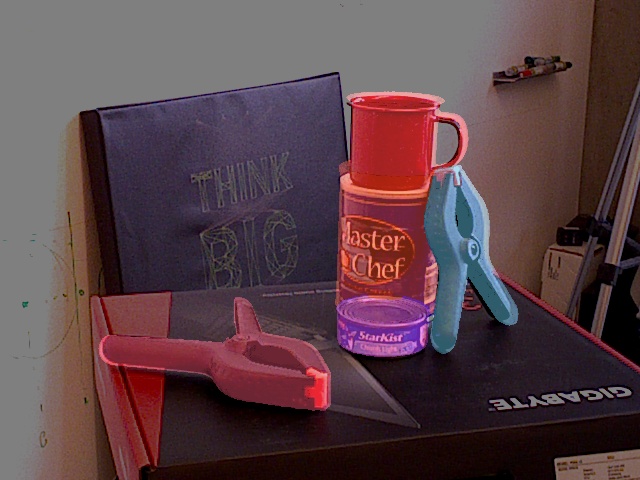}    
                        &   \includegraphics[width=\hsize,valign=m]{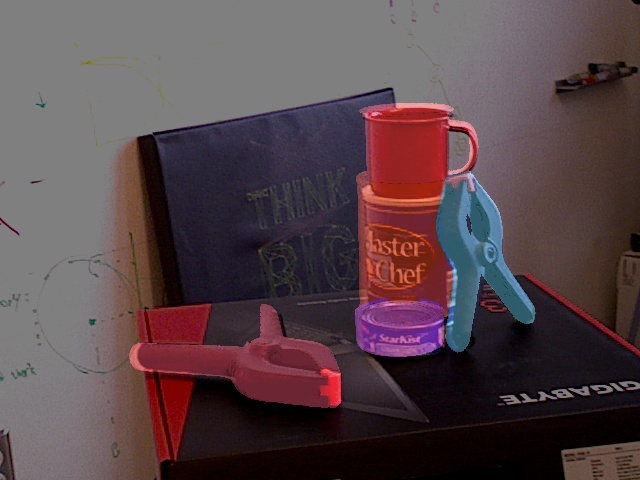}
                        &   \includegraphics[width=\hsize,valign=m]{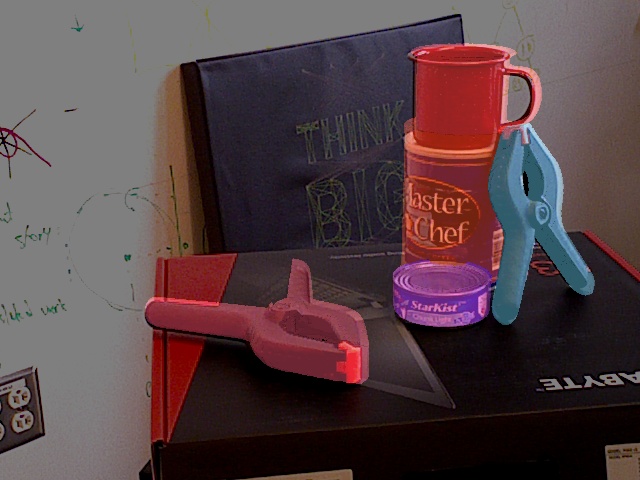}
                        &   \includegraphics[width=\hsize,valign=m]{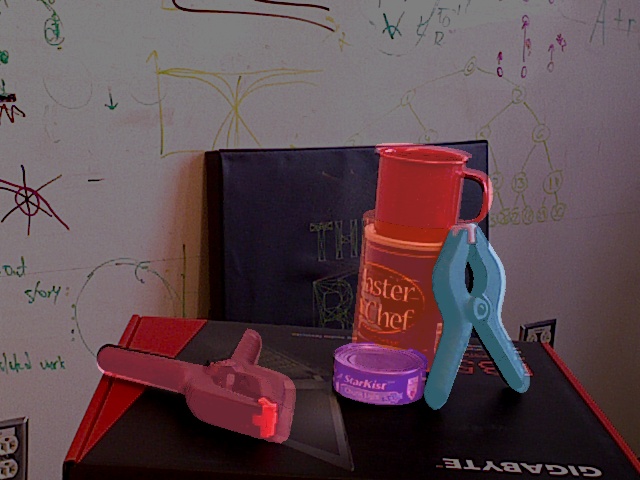}\\  \addlinespace[7pt]
                        
\rothead{CosyPose} &   \includegraphics[width=\hsize,valign=m]{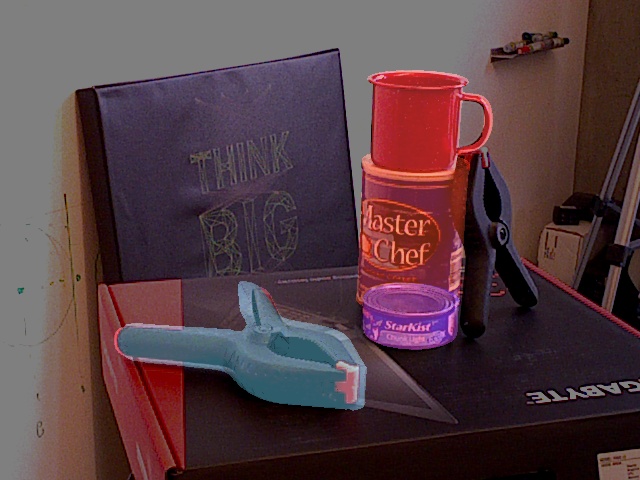}
                        &   \includegraphics[width=\hsize,valign=m]{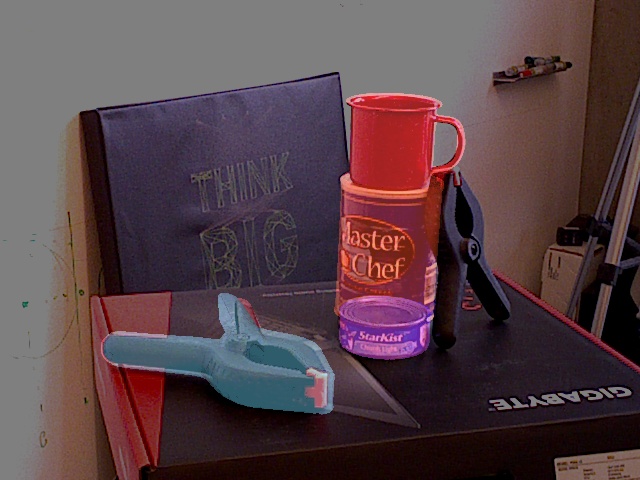}
                        &   \includegraphics[width=\hsize,valign=m]{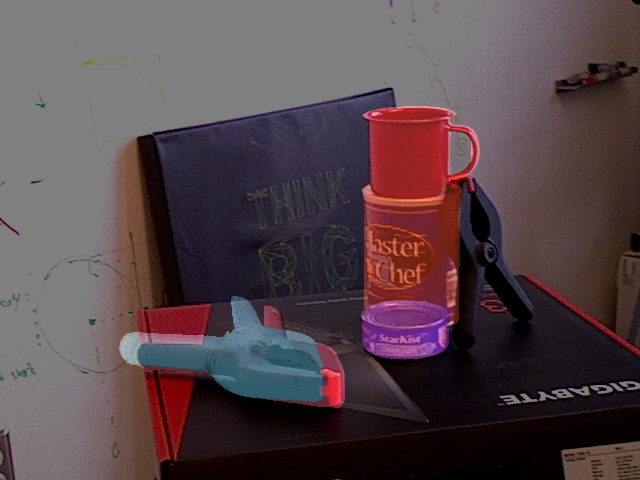}
                        &   \includegraphics[width=\hsize,valign=m]{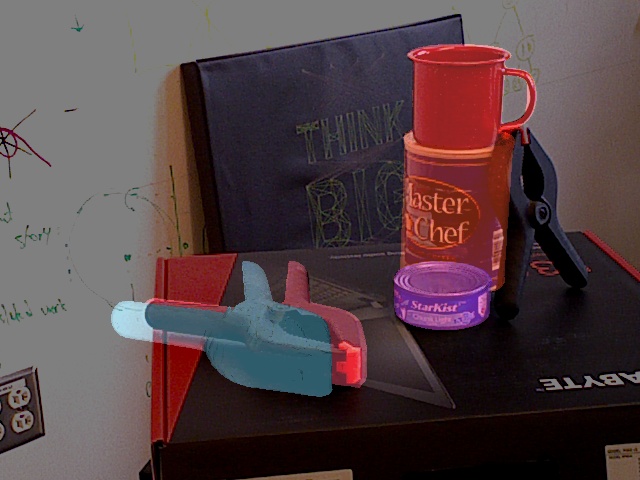}
                        &   \includegraphics[width=\hsize,valign=m]{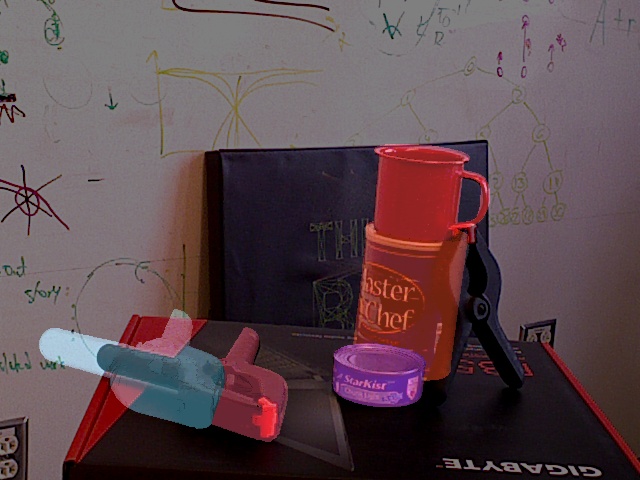}
                        \\  \addlinespace[7pt]

\rothead{Ours}        &   \includegraphics[width=\hsize,valign=m]{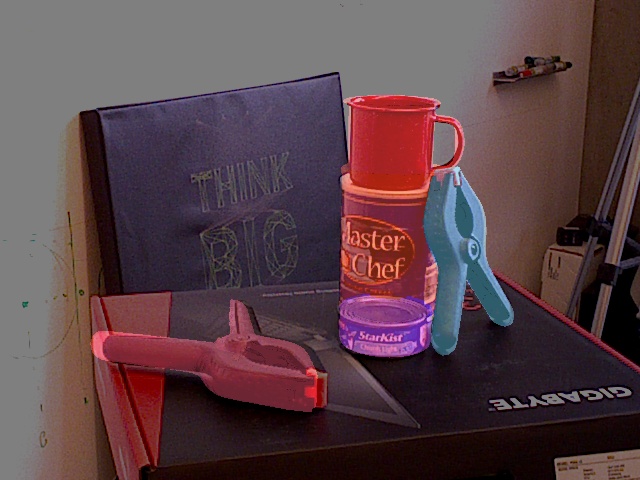}
                        &   \includegraphics[width=\hsize,valign=m]{figures/ycbv_dataset/27_001107_rtm}    
                        &   \includegraphics[width=\hsize,valign=m]{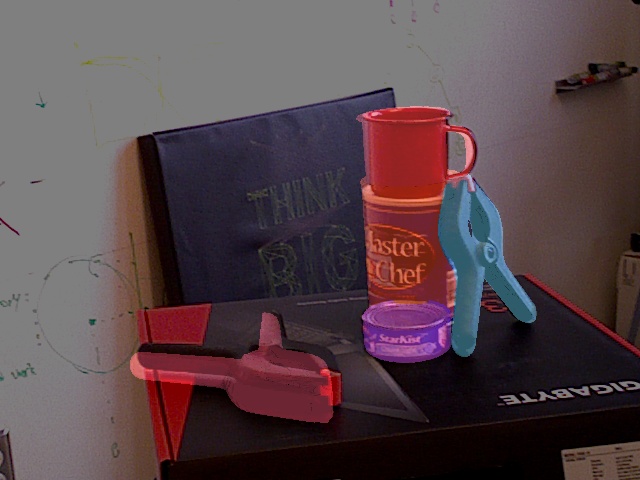}
                        &   \includegraphics[width=\hsize,valign=m]{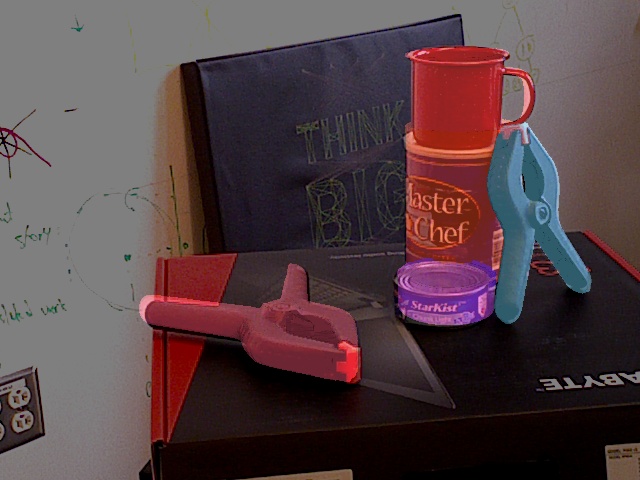}
                        &   \includegraphics[width=\hsize,valign=m]{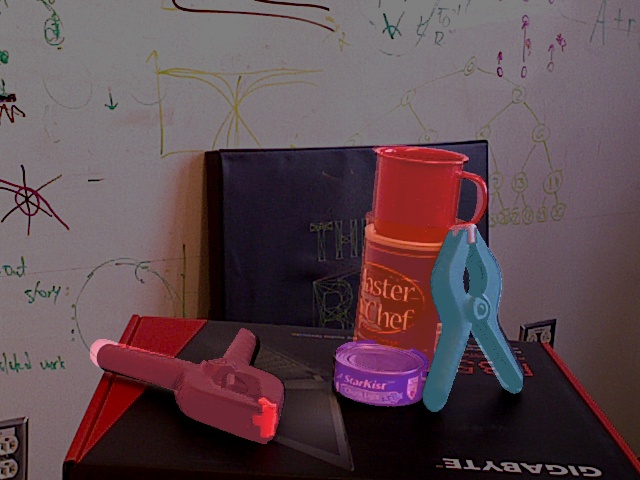}\\  \addlinespace[7pt]
                        
\end{tabularx}
    \caption{\textbf{Multi-view visualization of object poses in YCB-V dataset:} Our proposed approach, \approach, (bottom) is able to detect and optimize object poses in a more accurate way compared to the state-of-the-art CosyPose (center)~\cite{labbe2020cosypose} in multi-view setups. The ground truth is displayed on top. Differences are notable on the clamp right next to the can, and on the other clamp, a drift in different views becomes apparent.}
\label{fig:ycbv_multi}
\end{figure*}

\begin{figure*}[t!]
\setlength\tabcolsep{1pt}
\centering
\begin{tabularx}{\textwidth}{l XXXX }
& \centering Ground Truth & \centering YOLOPose + CosyPose & \centering CNOS + Megapose + CosyPose & \centering Ours \cr \\
\rothead{HoloLens (near)}        &   \includegraphics[width=\hsize,valign=m]{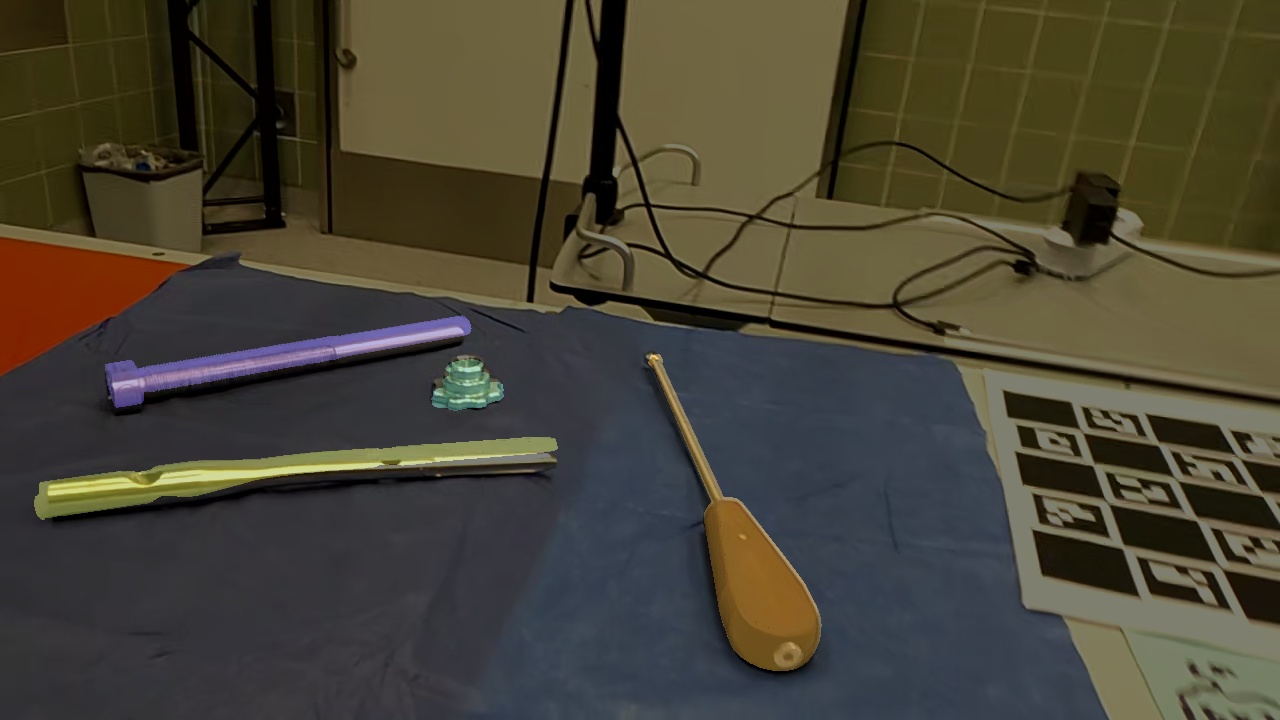}
                        &   \includegraphics[width=\hsize,valign=m]{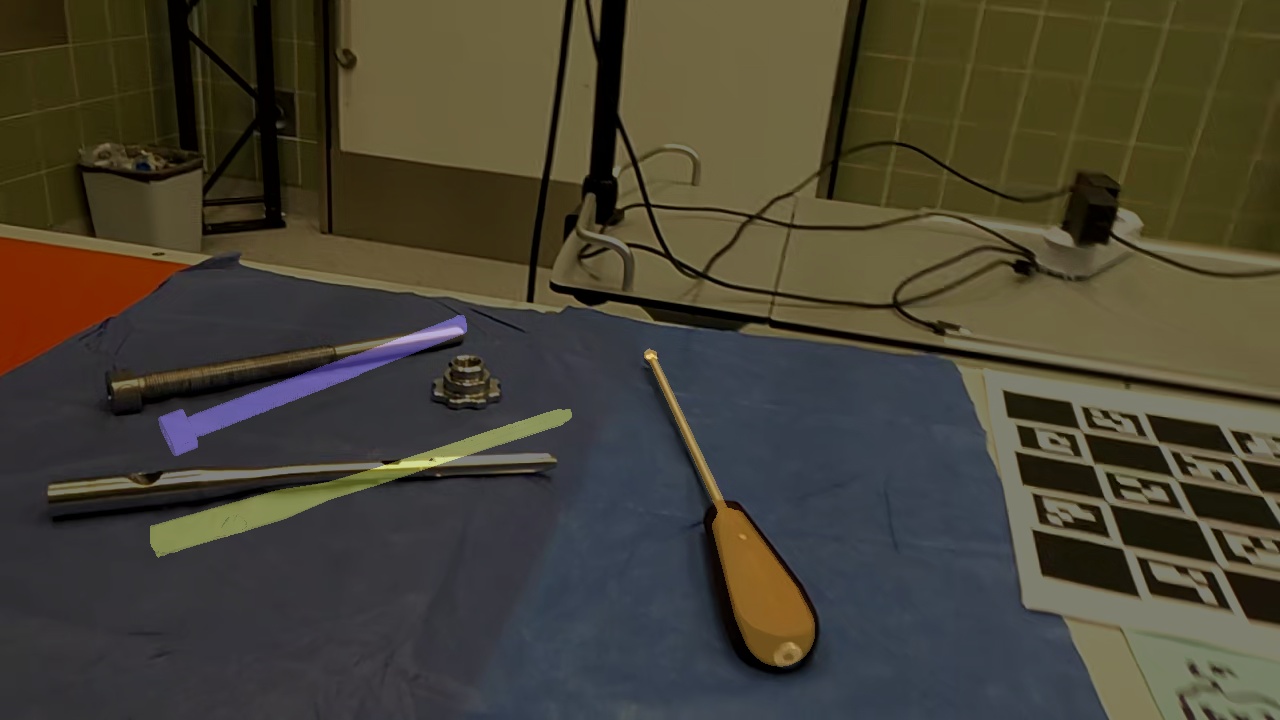}    
                        &   \includegraphics[width=\hsize,valign=m]{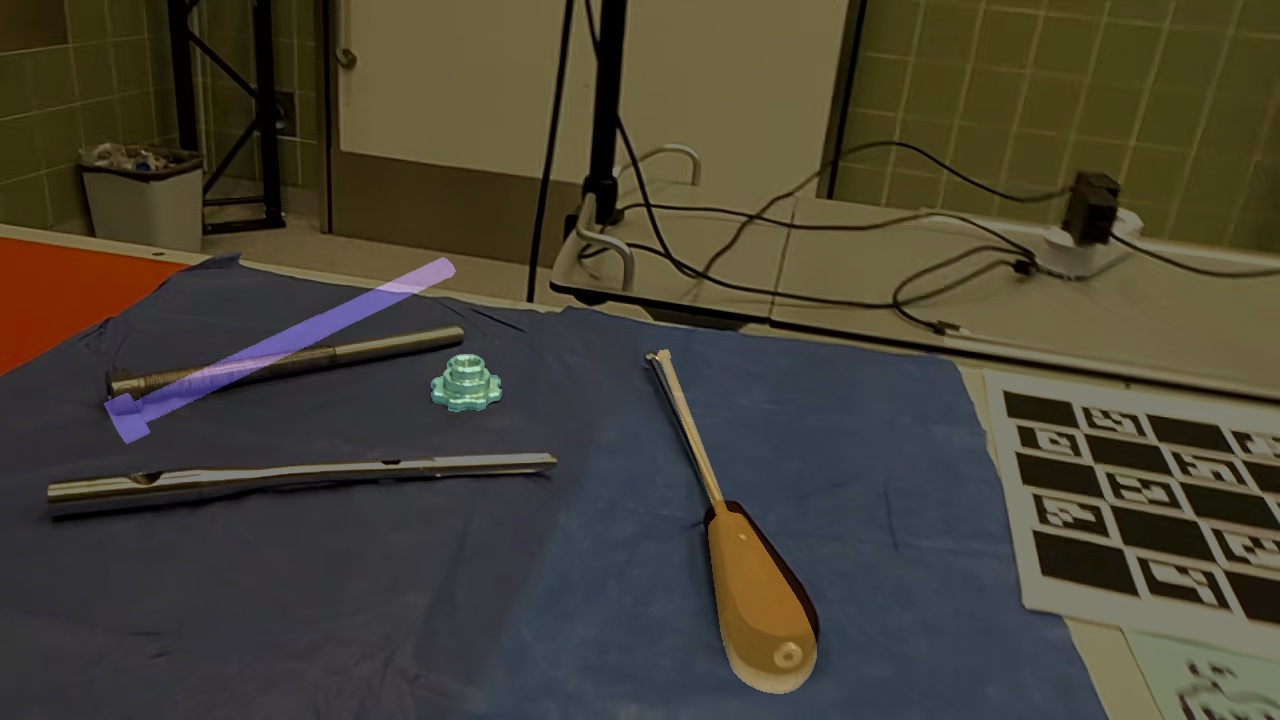}
                        &   \includegraphics[width=\hsize,valign=m]{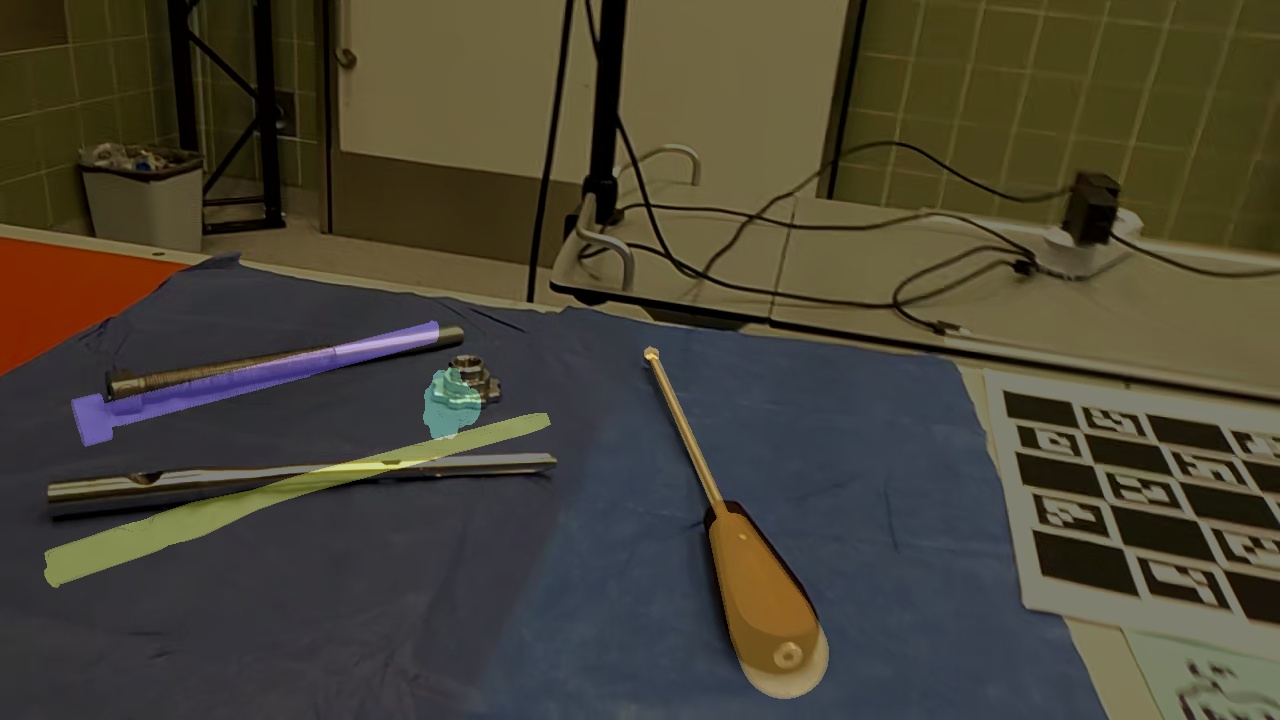}\\  \addlinespace[7pt]
                        
\rothead{Azure Kinect (near)} &   \includegraphics[width=\hsize,valign=m]{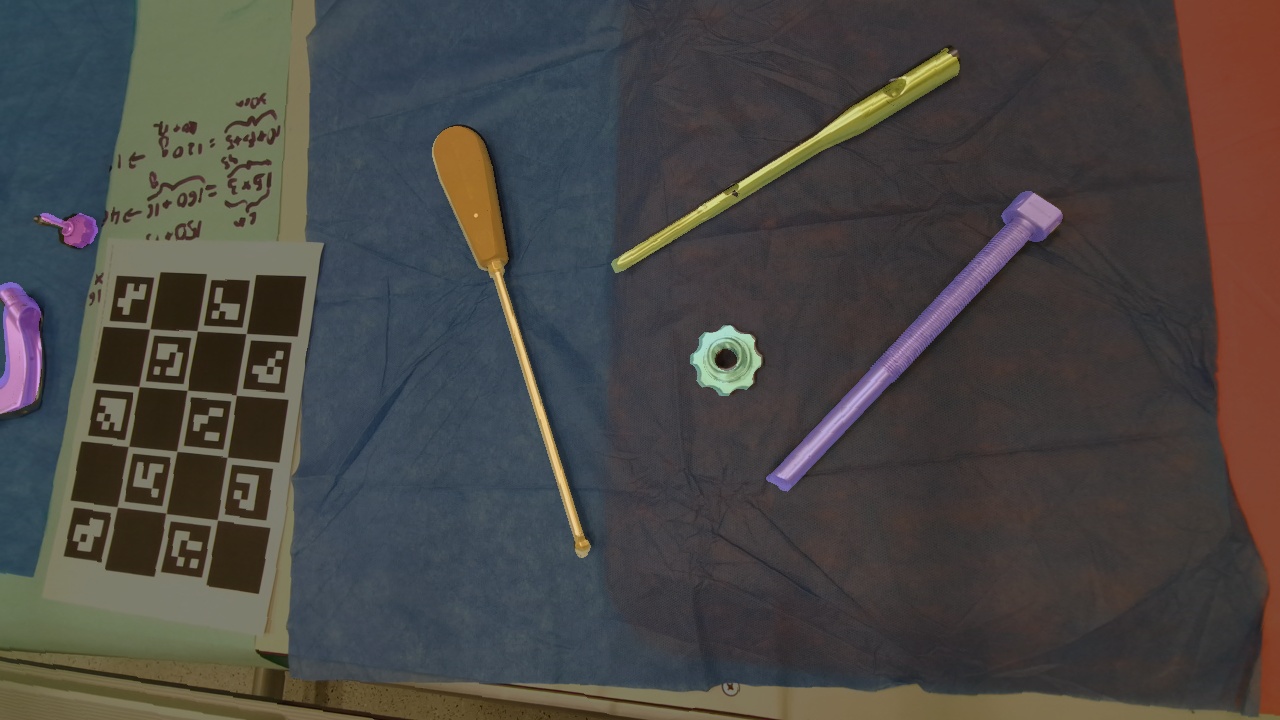}
                        &   \includegraphics[width=\hsize,valign=m]{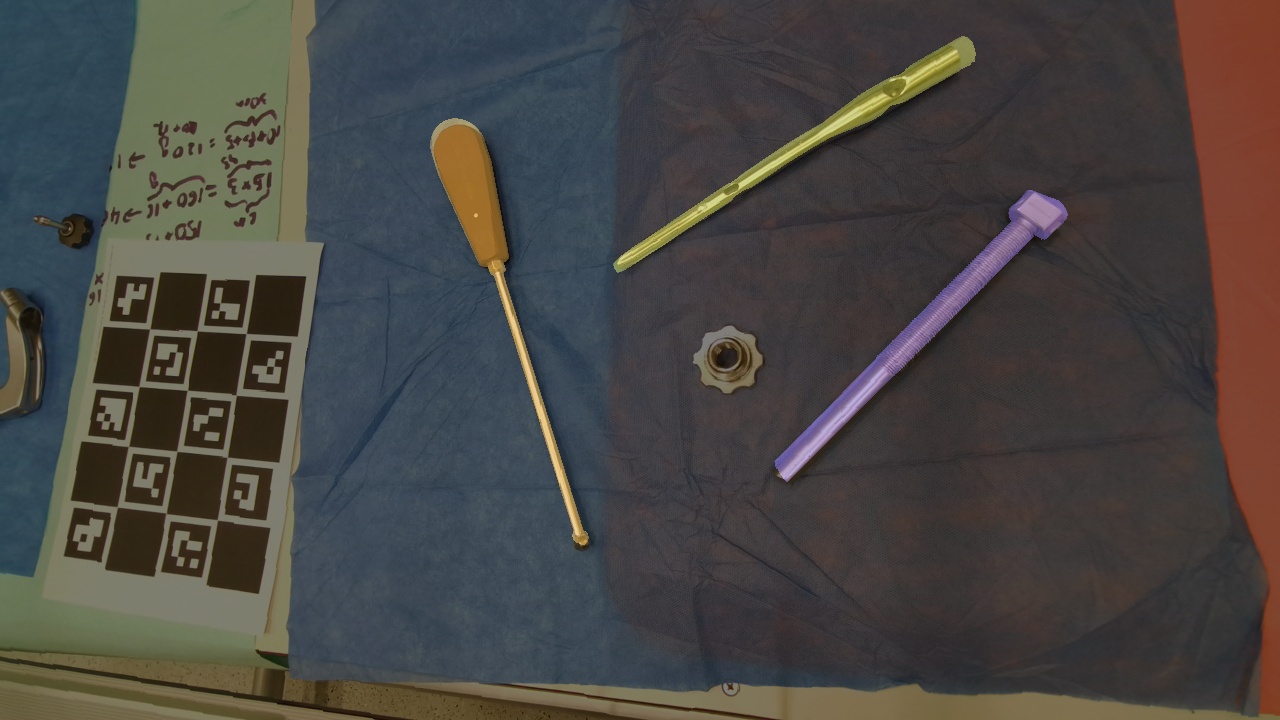}
                        &   \includegraphics[width=\hsize,valign=m]{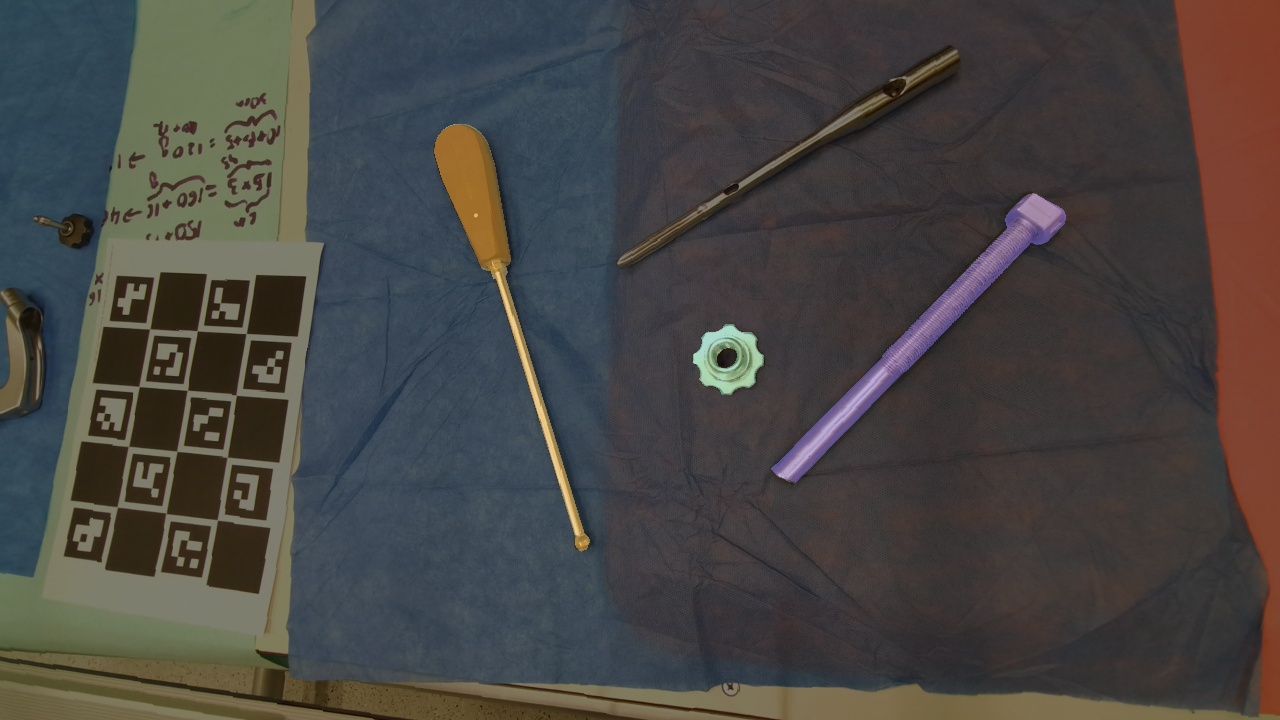}
                        &   \includegraphics[width=\hsize,valign=m]{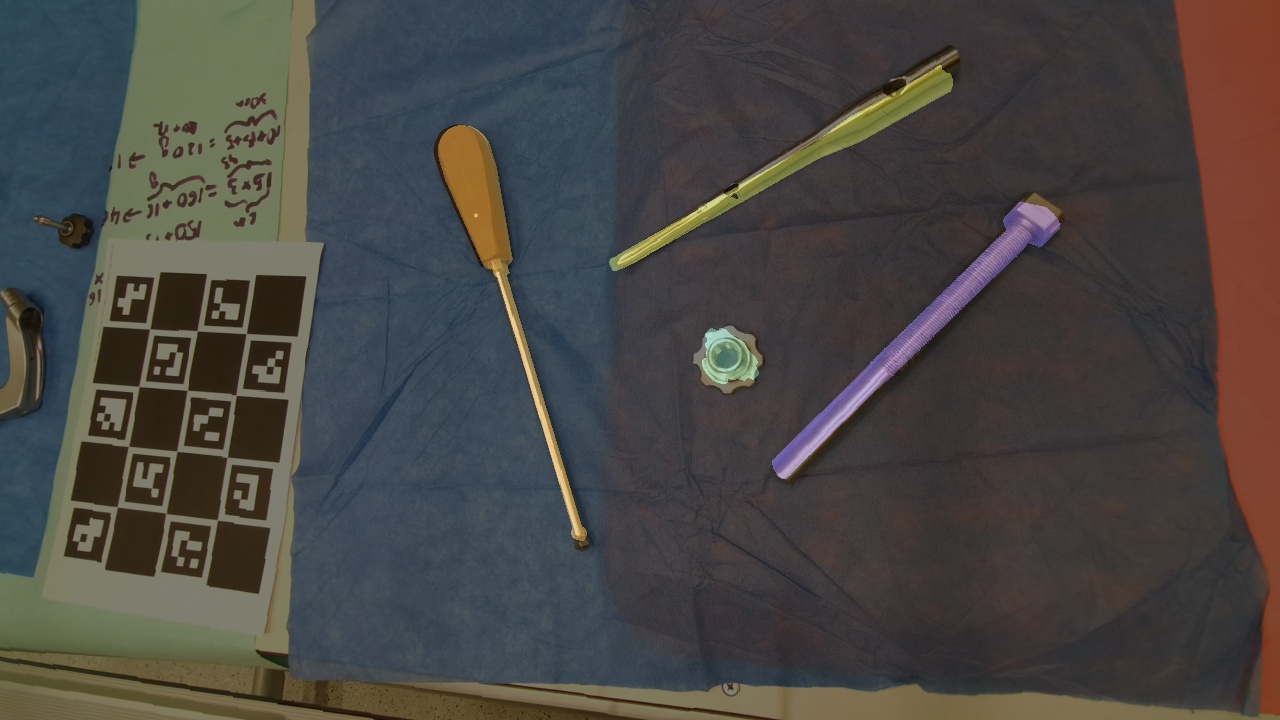}\\  \addlinespace[7pt]

\rothead{HoloLens (far)}        &   \includegraphics[width=\hsize,valign=m]{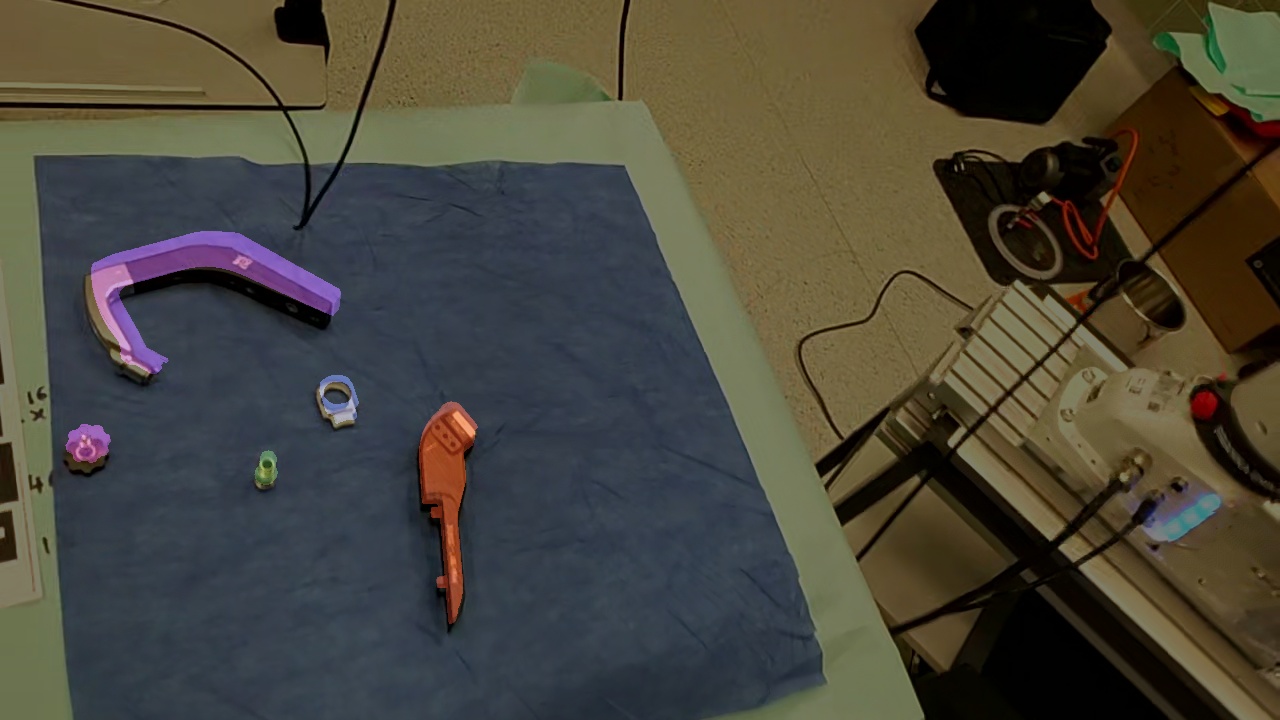}
                        &   \includegraphics[width=\hsize,valign=m]{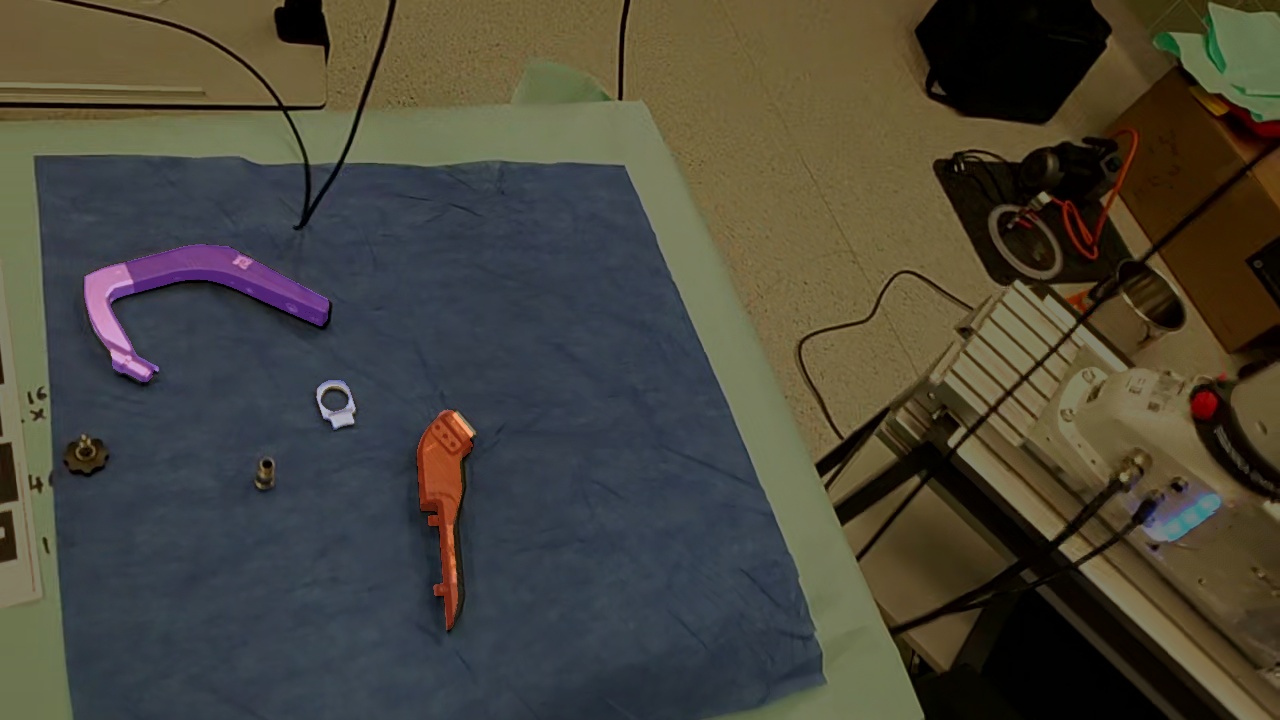}    
                        &   \includegraphics[width=\hsize,valign=m]{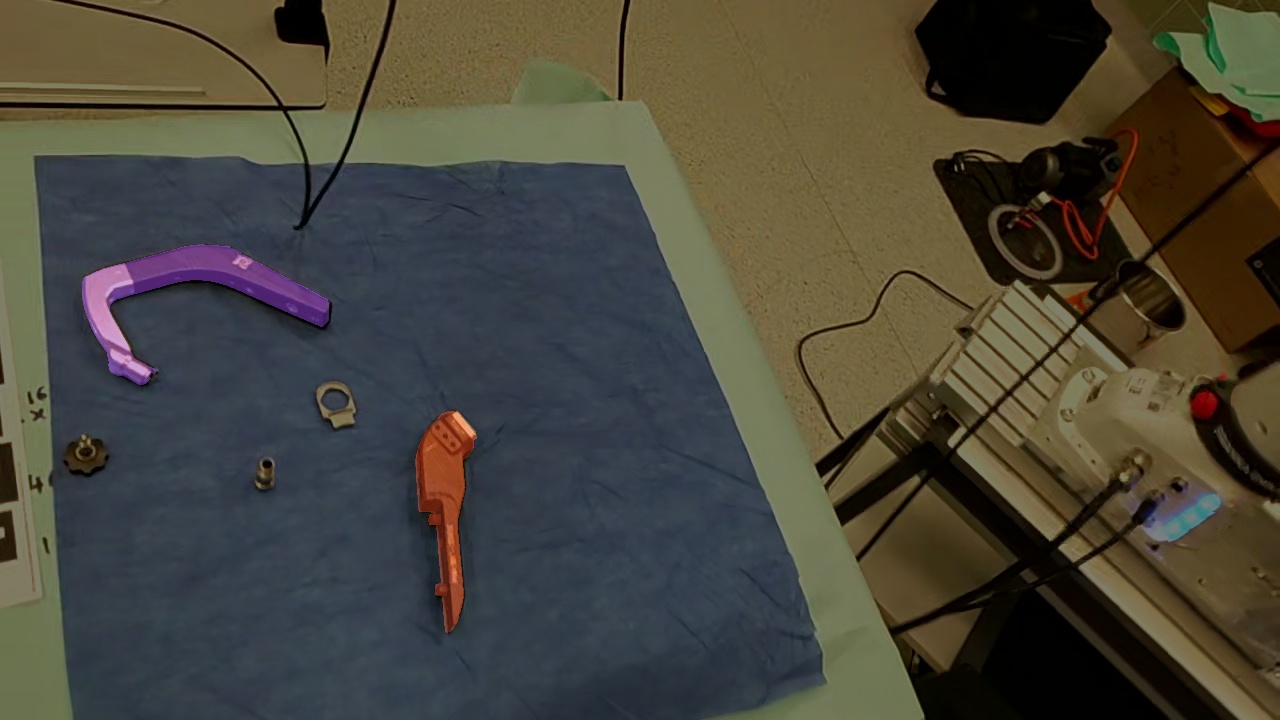}
                        &   \includegraphics[width=\hsize,valign=m]{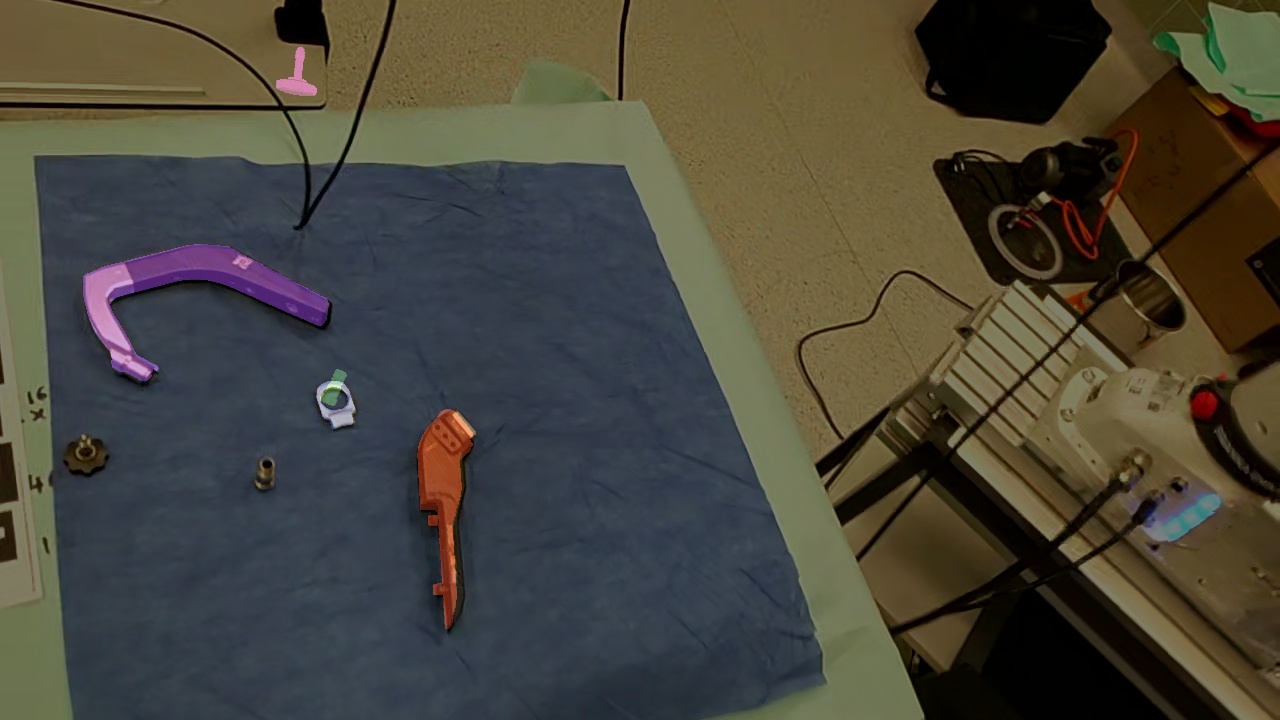}\\  \addlinespace[7pt]
                        
\rothead{Azure Kinect (far)} &   \includegraphics[width=\hsize,valign=m]{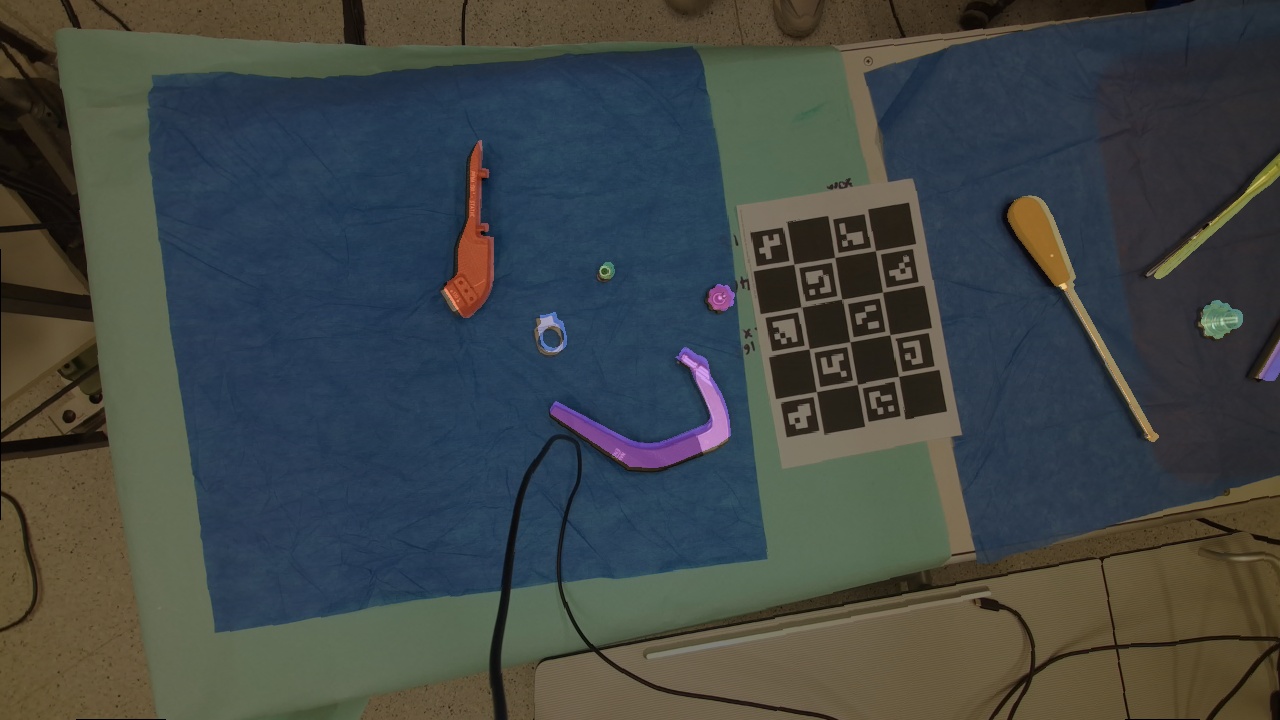}
                        &   \includegraphics[width=\hsize,valign=m]{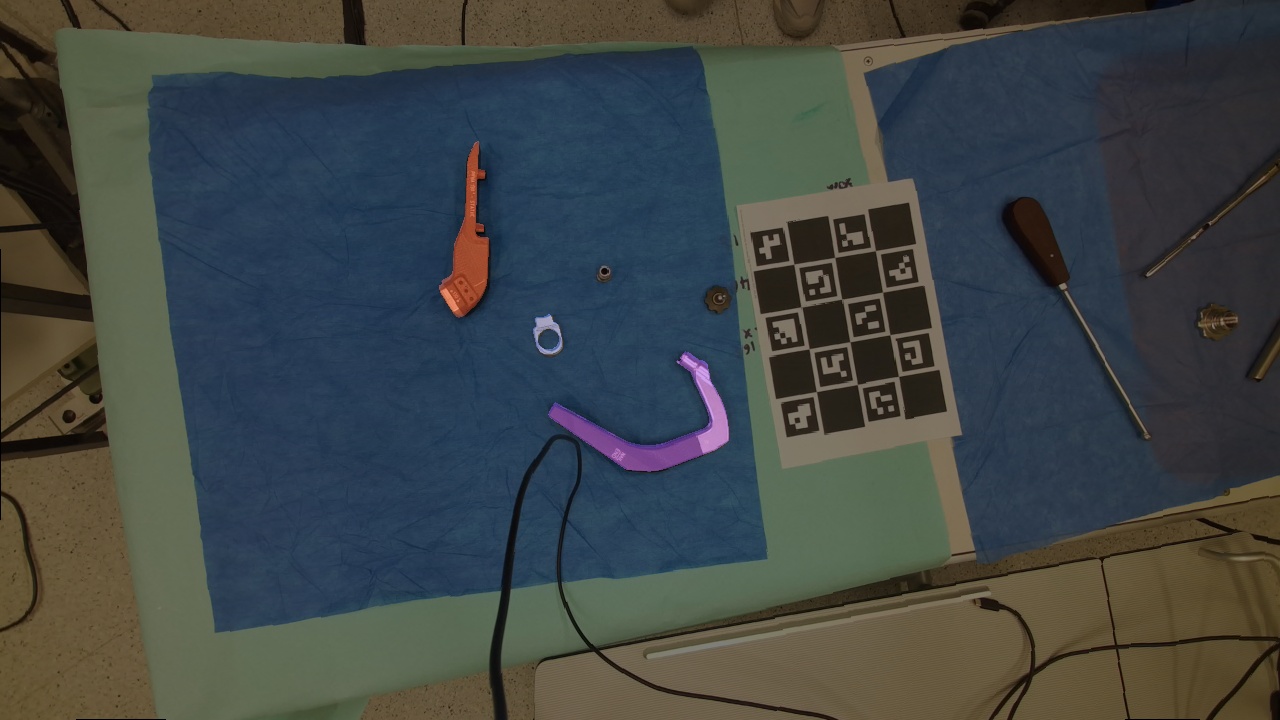}
                        &   \includegraphics[width=\hsize,valign=m]{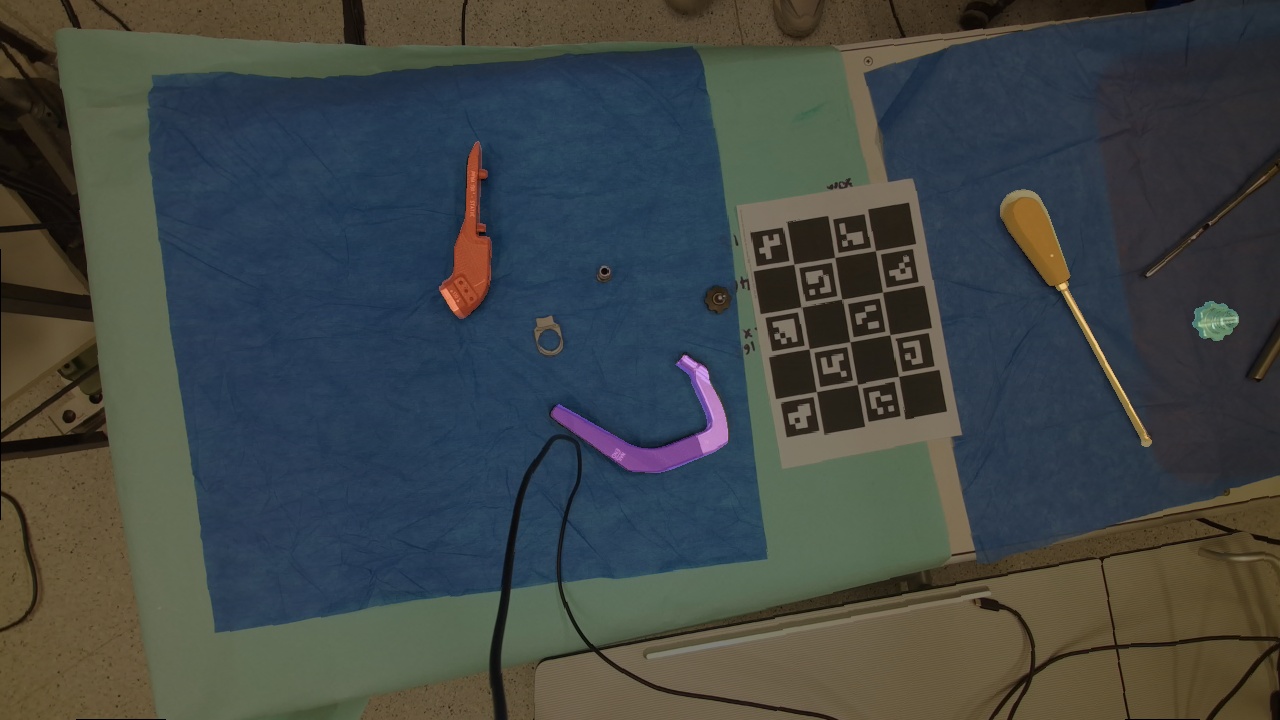}
                        &   \includegraphics[width=\hsize,valign=m]{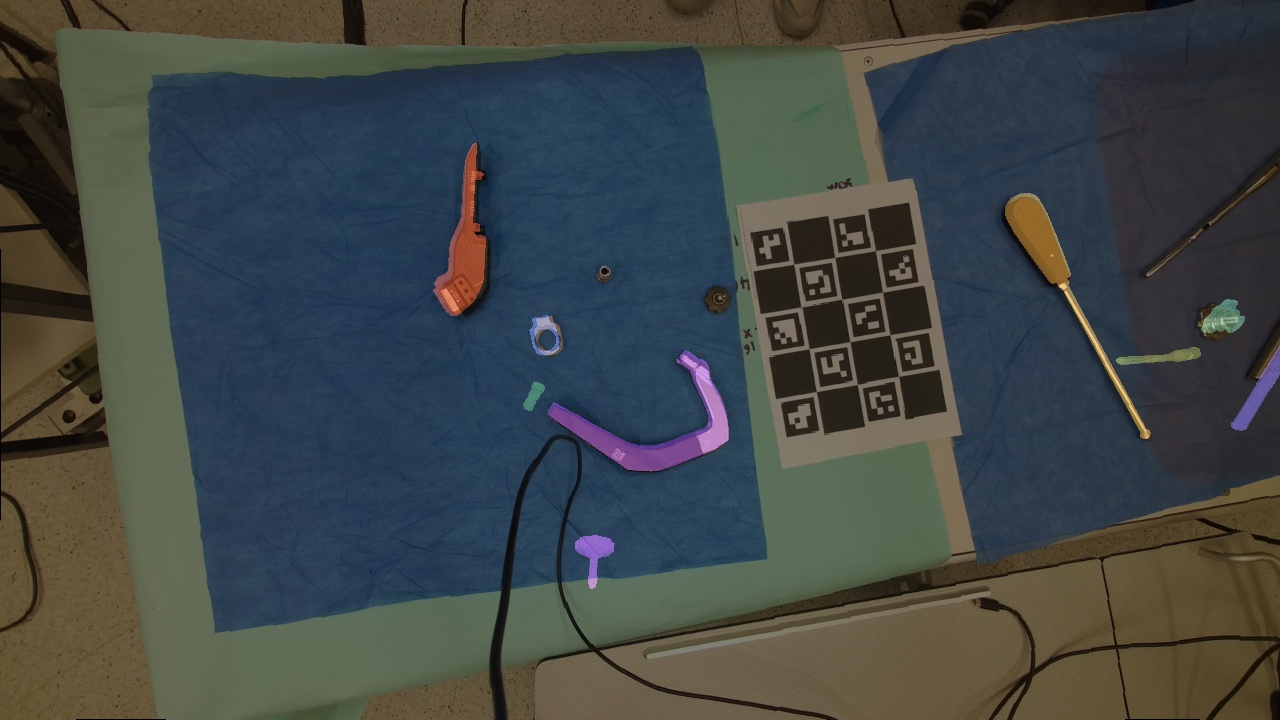}
\end{tabularx}
    \caption{\textbf{Multi-view visualization from our \dataset dataset in near distance (top two rows) and far distance (bottom two rows):} We compared YOLOPose + CosyPose \cite{labbe2020cosypose} (second left), CNOS \cite{nguyen2023cnos} + Megapose \cite{labbe2022megapose}  + CosyPose \cite{labbe2020cosypose} (second right), and our approach (\approach) (right) with the ground truth (left). Our approach has smaller visualization errors than the other approaches. \approach also integrates more objects using spatiotemporal information, leveraging shared and non-shared \ac{fov} information. Calibration markers presented in the scene are used to compare with our proposed marker-less approach.}
\label{tab:vis_near_far}
\end{figure*}

\begin{table}[!t]
    \centering
      \caption{\textbf{Quantitative multi-view 6D object pose results on YCB-V dataset~\cite{xiang_posecnn_2017}.} The best results per category (known vs. unknown camera poses) are printed in \textbf{bold}.  \label{tab:ycbv_mv}}
    \resizebox{\columnwidth}{!}{
    \begin{tabular}{l|c|ccc|ccc}
    \toprule 
    & Unknown &\multicolumn{3}{c|}{ADD-S AUC} & \multicolumn{3}{c}{ADD(-S) AUC} \\
    & Cam. Pose & 1 view &  3 views  &  5 views  & 1 view&  3 views  &  5 views  \\\midrule
    MV6D~\cite{duffhauss2022mv6d}              & \xmark & - &     91.2  &     91.1  & - &     85.6  &     84.0  \\
    SyMFM6D ~\cite{duffhauss2023symfm6d}          &\xmark &\textbf{96.8} & \textbf{95.4} & \textbf{95.4} & \textbf{94.1} & \textbf{91.7} & \textbf{91.6} \\ \midrule
    CosyPose~\cite{labbe2020cosypose}         & \cmark &89.8&     92.3  &     93.4  & 84.5 &     87.7  &     88.8  \\
    \approach(Ours)  &  \cmark     & \textbf{93.3} & \textbf{93.2} & \textbf{93.9} & \textbf{88.8} & \textbf{88.8} & \textbf{89.8} \\
    \bottomrule
    \end{tabular}}
\end{table}

\begin{table}[t]
    \caption{\textbf{Evaluation of camera pose estimation error before and after object-level bundle adjustment on the YCB-V dataset~\cite{xiang_posecnn_2017}:} We report pose
      rotation error (in mm) and translational error(in degrees) for the camera poses before and after bundle adjustment. Our proposed bundle adjustment helps improve both object and camera accuracy.}
    \label{tab:xyz_errors}
    \resizebox{\columnwidth}{!}{
    \begin{tabular}{l|c|cc|cc}
    \toprule 
      \multicolumn{1}{l|}{Methods} & Bundle  & \multicolumn{2}{c|}{3 views} & \multicolumn{2}{c}{5 views}\\
         & adjustment & $e_{trans} \downarrow$& $e_{ rot} \downarrow$ & $e_{trans} \downarrow$& $e_{ rot} \downarrow$\\
      \midrule
        CosyPose~\cite{labbe2020cosypose}   & \cmark & 22.30 & 1.49 & 27.57 & 1.90\\
        \approach (ours) & \xmark  & 35.04 & 2.39 & 36.85 & 2.52\\
        \approach (ours)  & \cmark  & \textbf{21.22} & \textbf{1.48} & \textbf{21.63} & \textbf{1.48}\\

      \bottomrule
    \end{tabular}}

\end{table}

\begin{table}[t!]
    \centering
    \caption{{\bf Effect of the 3D loss in YCB-V dataset~\cite{xiang_posecnn_2017}.} The proposed 3D loss outperforms the 2D key point loss. The larger the distance between the object and the camera is, the more obvious the advantage of the 3D loss.}
    \resizebox{\columnwidth}{!}{
    \begin{tabular}{lcc|c|c}
    \toprule
    Model & \ac{dcc} & 3D loss & ADD(-S) AUC & ADD-S AUC \\ \midrule
    Baseline & \xmark & \xmark & 63.9 & 77.4   \\
    Baseline & \cmark & \xmark & 80.7 & 88.2   \\
    Baseline & \cmark & \cmark& {\bf 83.4} & {\bf 89.4} \\ \midrule
    Delta & & & +20.5 & +12.0 \\
    \bottomrule
    \end{tabular} }
    
    \label{tab:error_adds}
\end{table} 

\begin{table}[t!]
    \caption{\textbf{Evaluation of camera pose estimation error on the T-LESS dataset~\cite{hodan_t-less_2017}.} We report pose
      rotation error (in mm) and translational error(in degrees) for the camera poses with state-of-the-art calibration and pose estimation methods.}
    \label{tab:tless_eval_cam}
    \resizebox{\columnwidth}{!}{
    \begin{tabular}{l|cc|cc}
    \toprule 
      \multicolumn{1}{l|}{Methods} & \multicolumn{2}{c|}{4 views} & \multicolumn{2}{c}{8 views}\\
         & $e_{trans} \downarrow$& $e_{ rot} \downarrow$ & $e_{trans} \downarrow$& $e_{ rot} \downarrow$\\
      \midrule
        CosyPose~\cite{labbe2020cosypose}    & 44.38 & 4.19 & 65.93 & 8.11\\
         ARToolKitPlus marker~\cite{wagner2007artoolkitplus} &64.27&14.01&61.97&13.48\\
         \approach(ours)   & \textbf{38.22} & \textbf{3.25} & \textbf{38.96} & \textbf{3.33} \\
      \bottomrule
    \end{tabular}}

\end{table}

\begin{table*}[t!]
\caption{\textbf{Camera pose estimation/Calibration results on our \dataset dataset:} The translation error is in millimeters (mm) ($e_{ave\_{trans}}$ denoted as $e_{trans}$, $\downarrow$), the rotation error ($e_{ave\_ rot}$ denoted as $e_{rot}$, $\downarrow$) is in degrees and the average runtime is in ms. The best results among all methods are labeled in bold.}
    \label {tab:op_quan_eval}
\centering
   \resizebox{\textwidth}{!}{    
        \begin{tabular}{l|ccc|ccc|ccc|ccc}\toprule
        \multicolumn{1}{l|}{Methods} & 
        \multicolumn{9}{c|}{{\centering Object-level}}&
        \multicolumn{3}{c}{{\centering Calibration board}}\\ \midrule
        
        \multirow{2}{*}{Conditions} 
        &\multicolumn{3}{c|}{YOLOpose + CosyPose~\cite{labbe2020cosypose}} &\multicolumn{3}{c|}{CNOS~\cite{nguyen2023cnos} + Megapose~\cite{labbe2022megapose} + CosyPose~\cite{labbe2020cosypose}} & \multicolumn{3}{c|}{\approach(ours)} & \multicolumn{3}{c}{Charuco marker}\\ 
         & $e_{trans} \downarrow$  & $e_{ rot} \downarrow$ & runtime  
         & $e_{trans} \downarrow$  & $e_{ rot} \downarrow$ & runtime 
         & $e_{trans} \downarrow$  & $e_{ rot} \downarrow$ & runtime 
         & $e_{trans} \downarrow$  & $e_{ rot} \downarrow$ & runtime\\ \midrule
         Near       & 64.54 & 9.81 & 135.05 & 47.17 & 6.84 & 84831.93  & 45.54 & 6.48 & 43.37 & \textbf{34.98} & \textbf{3.40} & \textbf{21.72}\\
         Far      & 94.69 & 8.90 & 194.70 & 70.47 & 6.91 & 79163.96  & \textbf{52.79} & \textbf{5.53} & 47.65 & 81.68 & 6.76 & \textbf{18.56}\\  \midrule 
         Mean    & 79.62 & 9.36 & 164.9 & 58.8 & 6.88 & 81997.95 & \textbf{49.17} & 6.01 & 45.51 &  58.33 & \textbf{5.08} & \textbf{20.14}\\
        \bottomrule
        \end{tabular}}
\end{table*}

\begin{table}[t!]
    \centering
    \caption{
        \label{tab:op_obj}
        \textbf {Multiview object pose evaluation on \dataset ADD(-S)-AUC, translation error in mm and rotation error in degrees with near and far conditions.} Symmetric objects are denoted by $\star$. Rotational errors are only evaluated for unsymmetrical objects.}
       
    \resizebox{\columnwidth}{!}{
    \begin{tabular}{l | c c  c | c c c } \toprule
        \multirow{2}{*}{Methods} &\multicolumn{3}{c|}{YOLOpose + CosyPose~\cite{labbe2020cosypose}} &        \multicolumn{3}{c}{\approach(ours)} \\
        & $ADD(S) AUC\uparrow$ & $e_{trans} \downarrow$ & $e_{ rot} \downarrow$& $ADD(S) AUC\uparrow$& $e_{trans} \downarrow$ & $e_{ rot} \downarrow$\\ \midrule
        Aiming arm        & 0.00 & 158.62 & 20.53  & 30.90 & 85.57 & 19.00\\
        Bar$\star$               & 0.00 & 244.59 &  - & 37.51 & 109.23 & - \\
        Gear$\star$              & 33.33 & 156.18 & -  & 47.65 & 127.31 & - \\
        Insertion handle  & 0.00 & 215.67 & 13.01  & 56.29 & 46.62 & 9.13\\
        Nail              & 0.00 & 261.23 & 140.67  & 46.59 & 75.46 & 83.61\\
        Roller$\star$            & 0.00 & - & -  & 26.32 & 289.95 & -\\
        Screw$\star$             & 0.00 & - & -  & 5.04 & 206.71 & -\\
        Spacer            & 0.00 & 196.19 & 15.48  & 39.82 & 99.69 & 69.58\\
        Stick$\star$             & 20.00 & 197.66 &  - & 27.12 & 195.27 & -\\
        MEAN              & 5.93 & 158.91 &  47.43 & \textbf{35.25} & \textbf{137.31} & \textbf{45.33}\\ \bottomrule
    \end{tabular}
    }  
\end{table}


        
        
    

To evaluate our proposed multiple-view approach, we also compared it with CosyPose~\cite{labbe2020cosypose}, MV6D~\cite{duffhauss2022mv6d}, and SyMFM6D~\cite{duffhauss2023symfm6d}. \autoref{tab:ycbv_mv} shows our quantitative evaluation results. Our methods are better than CosyPose and MV6D in all views, with ADD(S) AUC around 88 and ADD-S AUC around 93, 
while SyMFM6D achieves ADD-S AUC around 95 - 96 only with known camera poses. We can observe that the object pose accuracy increases with camera view numbers, which shows similar results to CosyPose. Our results with multiple views show slightly better results than our experiments on single view. This is because the YCB-V dataset has few camera poses variety which may not significantly contribute to object pose accuracy. \autoref{fig:ycbv_multi} shows the multi-view pose visualization of our approach compared with CosyPose. Our approach shows better object and pose detection accuracy.

\begin{figure}[t!]
    \centering
    {{\includegraphics[width=\columnwidth]{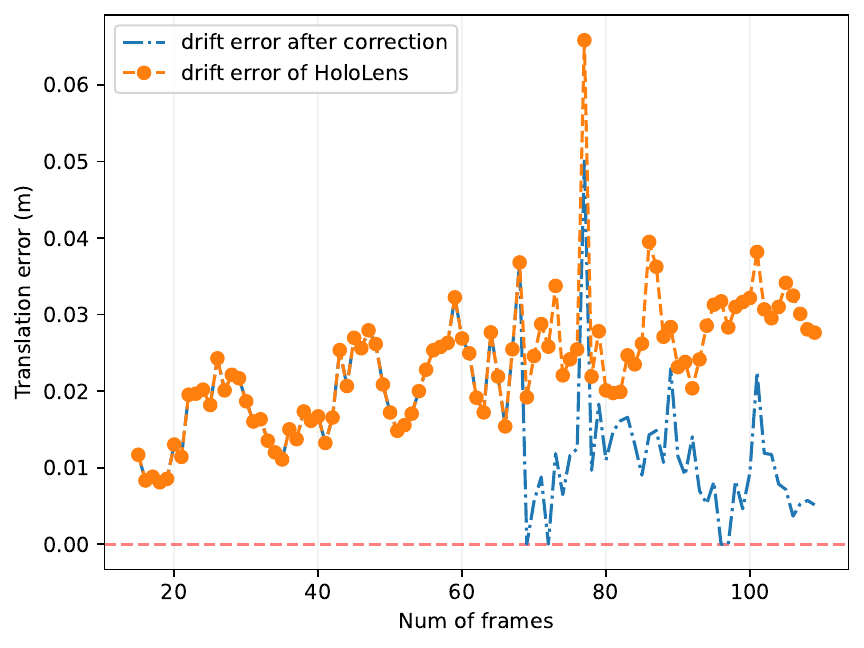} }}%
    \caption{Drift error correction with \approach: The figure shows the increasing sensor drift (orange) which can be reduced when on-the-fly camera pose estimation is performed (blue).}%
    \label{fig: drift_error}%
\end{figure}

\autoref{tab:xyz_errors} shows evaluation results of camera poses before and after our proposed object-level bundle adjustment. As seen in  \autoref{tab:xyz_errors}, there is an improvement in camera pose accuracy with 3 views. Before the bundle adjustment, the average translation error is 35.04 mm, and the rotation error is 2.39 degrees. After the camera pose optimization, we can observe the lower translation error of 21.22 mm and rotation error of 1.48 degrees. We observed that adding additional views does not further improve the camera pose estimation in terms of translation and rotation errors.

We tested each module individually to show its individual effects on the overall performance. As shown in \autoref{tab:error_adds}, the \ac{dcc} module improves the accuracy for single-stage 6D pose estimation with more than 10\% in ADD(-S) AUC and ADD-S AUC. Introducing 3D keypoint loss helps increase ADD(-S) AUC by 2.7 and ADD-S AUC by 1.2. Our proposed new module has improved the performance of single-view 6D object pose estimation, which is the base of the multi-view approach.

To study the effects of symmetric and texture-less objects, we also evaluate the camera pose estimation and calibration error on the T-LESS dataset, as shown in \autoref{tab:tless_eval_cam}. The ARToolKitPlus marker~\cite{wagner2007artoolkitplus} has an overall larger error than the marker-less pose estimation approach. Our method is better than the state-of-the-art methods with a translation error of around 40mm and a rotation error of 3 degrees.

\subsection{Evaluation on \dataset dataset}

\autoref{tab:op_quan_eval} shows the quantitative comparison of camera pose estimation methods and calibration with markers. In the condition of near distance, the accuracy of the marker with 34.98 mm and 3.4 degrees is better than our object-level pose estimation with 45.54 mm and 6.48 degrees. In addition, our object-level methods outperform other methods both in speed and accuracy in this condition.

In the condition of far distance, our scene matching method, with an accuracy of 52.79 mm and 5.53 degrees, outperforms both the other object-level approach and the marker-based approach. The calibration board performs worse than the near distance condition, with 81.68 mm and 6.76 degrees. In both conditions, our runtime with around 45 ms is smaller than the other state-of-the-art multi-view pose estimation approaches, but larger than the marker-based approach with around 20 ms.  
Our approach outperforms the marker-based approach in the condition of far distance and proves its suitability for marker-less \ac{ar} applications.

\autoref{tab:op_obj} shows the quantitative evaluation of object poses. As shown in  \autoref{tab:op_obj}, small objects like a roller and a screw have a large offset of more than 200 mm. Large objects, like the insertion handle with an error of 46.64 mm, perform relatively better than these small objects. In addition, the average object pose error is 137.31 mm and 45.33 degrees, while the camera pose error is only around 50 mm, as shown in \autoref{tab:op_quan_eval}. The error of camera pose is roughly the pose error of the best object pose candidates. As a result, the camera poses  have better accuracy than object poses, which means our approach can effectively calculate camera poses robustly with object inliers. The object poses of the roller and the screw are not available for the state-of-the-art approach CosyPose, as it removes these object outliers, while we keep these objects. We show overall better results than previous methods with our proposed spatiotemporal reasoning and pose optimization.

\autoref{tab:vis_near_far} shows the visualization comparisons of object poses in different camera views with near and far distances. As shown in the top two rows of the figure with near condition, our approach is able to detect more objects than the other approaches using spatiotemporal information. In addition, our approach has fewer visualization errors than other methods. As shown in the bottom two rows of the figure with the far condition, our methods are able to efficiently fuse object information in different camera views and recover the camera poses robustly. Our dynamic scene graph, which integrates HoloLens, can keep most of the object poses, which works better than the other static pose estimation approaches. However, some small objects have large offsets due to errors in the network.

\autoref{fig: drift_error} shows the drift error corrections with our proposed on-the-fly pose estimation methods. As shown in the figure, there are increasing drift errors between the HoloLens \ac{slam} and ground truth data. Our proposed method can reduce the drift offset between \ac{hmd} and external cameras by updating poses. The drift error is zero in the time stamp with updated pose estimation and accumulates over times in other frames.

\section{Discussion}

We propose \approach to enable markerless and continuous real-time multi-view camera pose estimation of both static and dynamic cameras using known objects. We use spatiotemporal \ac{fov} overlaps of known objects and enhance state-of-the-art pose estimation algorithms to update a spatiotemporal scene graph. By including dynamic cameras, our approach makes pose estimation of a non-overlapping multiple-camera set-up possible.


 

\subsection{Accuracy and Pose Error Analysis}
Assessing the accuracy of object and (in turn) camera pose estimation revealed a two-fold outcome. On the one hand, our object pose estimator has been improved compared with previous approaches regarding the single-camera results, as shown in \autoref{tab:ycb_results_ADD}. On the other hand, our multi-camera method also showed better results with the same single-camera inputs than previous methods, as shown in \autoref{tab:op_quan_eval} and \autoref{tab:op_obj}. To evaluate the effects of multi-camera methods, we use the same single-camera input with YOLOpose for Cosypose and our proposed approach in our new contributed \dataset dataset. Our approach shows better results than the current state-of-the-art object-level pose estimation approaches in both conditions, with 45.54 mm, 6.48 degrees error in near distance and 52.79 mm, 5.53 degrees error in far distance.

Pose errors depend on many factors, such as object/marker size, distance to the camera, rolling-shutter artifacts or motion blur of dynamic cameras. Especially in our proposed dataset, the errors are large for small objects due to a lack of features, which makes the detection difficult. Both \autoref{tab:ycb_results_ADD} and \autoref{tab:op_obj} show that the pose error varies for different objects. Our proposed methods can calculate camera pose best using reliable overlapping object poses based on multiple candidates. By choosing a set of most robust and precise object candidates, the estimated camera poses can be even more accurate than the object poses. Our methods also show better results in the cluttered scenes of the T-LESS dataset, where markers can be occluded. The occlusion-aware object pose estimation models can overcome these challenges, while marker-based calibration is error-prone in these scenarios. Regarding to errors from dynamic cameras, rolling-shutter artifacts or motion blur may happen in frames with high-speed camera motions. We reduce these influences by selecting key frames with suitable threshold.

Our evaluation on real data shows results similar to marker-based calibration methods. Our proposed marker-less approach even outperforms the ARToolKitPlus marker in the T-LESS dataset, as shown in \autoref{tab:tless_eval_cam}, and Charuco markers in the far condition of our \dataset dataset, as shown in \autoref{tab:op_quan_eval}. We assume that the calibration quality depends highly on the distance to the calibration board. Our approach filters object outliers and relies on multiple objects instead of a single, smaller calibration marker. Thus, our approach can better recover object and camera poses in the far condition.


\subsection{Runtime and Latency}

The \approach pipeline runs at approx. 50ms (20 FPS) for three views using temporal overlapping keyframes in our dataset. Our real-time performance is achieved, by exploiting spatiotemporal continuity in multi-view video sequences. The pose estimation time is around 40 ms and candidate matching time is around 10 ms for each frame. The object-level bundle adjustment is only applied for the overlapping frame with threshold and the runtime is around 50ms for these overlapping keyframes, but less than 1 ms in average for all frames. We assume that this is applicable as a baseline for most \ac{ar} applications' multi-view sensing. However, future work can finetune parameters to individually balance speed and accuracy, specifically for each use case and available computational power.

To account for different camera latencies and moving objects, we only consider objects in a consistent spatial layout from each camera for pose estimation. Thus, we exclude moving objects which can not be used for pose estimation. Our methods do not require strict synchronized multiple cameras, which also explain the idea of spatiotemporal pose estimation, which is performed in different time and space. The essence of the task is to align multiple cameras in the same coordinate system and these cameras may not be in the same time stamp. Given continuous SLAM sensing on the dynamic camera, the individual camera latency does not impede the pose estimation process. It will only change the time when the pose estimation happens.


\subsection{Limitations}

The current system runs in real time, but is limited to known objects (state-of-the-art instance-level pose estimator). Generalizable pose estimators like Megapose \cite{labbe2022megapose} and object detectors like CNOS ~\cite{nguyen2023cnos} can overcome this limitation, but still suffer from high computational time. Up-scaling this and getting towards unknown objects can not match the real-time requirement. To make the overall pipeline work for novel objects, an appropriate real-time object pose estimator will be required.

\section{Conclusions}

We enable multi-view camera pose estimation for \ac{ar} applications using static and dynamic cameras. \approach is a object and camera pose estimation toolkit. Our approach, \approach, does not require dedicated optical markers, but uses known object in a temporarily overlapping \acp{fov}, and continuously updates camera poses in real-time. We propose a dynamic scene graph for joint estimation of cameras' and objects' poses using object-level bundle adjustment. Since there is no suitable dataset for this scenario, we also present a novel dataset using two static and one dynamic camera. We achieve a runtime of 45.51 ms in our \dataset dataset with three camera views using keyframes with overlapping \ac{fov} and even outperform a marker-based approach in the far condition. Moreover, we outperform current state-of-the-art pose estimators in the YCB-V, T-LESS, and our \dataset datasets while balancing speed and accuracy. 

Our work is a promising step towards a markerless multi-camera and multi-object pose estimation for \ac{ar} applications to replace the current widely used marker-based approaches. Future work can build on our baseline and enhance the precision and runtime (e.g., by combining object pose estimation and tracking). 
\acknowledgments{%
This work was funded by the German Federal Ministry of Research, Technology and Space (BMFTR) in the grant program "AI-based assistance systems for in-process health applications", grant number 16SV8973. Shiyu Li acknowledges the financial support from the China Scholarship Council. %
}

\bibliographystyle{abbrv-doi-hyperref}

\bibliography{template}

\appendix 

\section{About Appendices}
Refer to \cref{sec:appendices_inst} for instructions regarding appendices.

\section{Troubleshooting}
\label{appendix:troubleshooting}

\subsection{ifpdf error}

If you receive compilation errors along the lines of \texttt{Package ifpdf Error: Name clash, \textbackslash ifpdf is already defined} then please add a new line \verb|\let\ifpdf\relax| right after the \verb|\documentclass[journal]{vgtc}| call.
Note that your error is due to packages you use that define \verb|\ifpdf| which is obsolete (the result is that \verb|\ifpdf| is defined twice); these packages should be changed to use \verb|ifpdf| package instead.

\subsection{\texttt{pdfendlink} error}

Occasionally (for some \LaTeX\ distributions) this hyper-linked bib\TeX\ style may lead to \textbf{compilation errors} (\texttt{pdfendlink ended up in different nesting level ...}) if a reference entry is broken across two pages (due to a bug in \verb|hyperref|).
In this case, make sure you have the latest version of the \verb|hyperref| package (i.e.\ update your \LaTeX\ installation/packages) or, alternatively, revert back to \verb|\bibliographystyle{abbrv-doi}| (at the expense of removing hyperlinks from the bibliography) and try \verb|\bibliographystyle{abbrv-doi-hyperref}| again after some more editing.

\end{document}